\documentclass{article} 

\usepackage{iclr2027_conference,times}
\usepackage{enumitem}

\usepackage{amsmath,amsfonts,bm}

\def\eqref#1{equation~\ref{#1}}

\def\1{\bm{1}}

\DeclareMathAlphabet{\mathsfit}{\encodingdefault}{\sfdefault}{m}{sl}
\SetMathAlphabet{\mathsfit}{bold}{\encodingdefault}{\sfdefault}{bx}{n}

\usepackage{amsmath,amssymb,bm}
\usepackage{microtype}
\usepackage{enumitem}
\usepackage{hyperref}
\usepackage{url}
\usepackage{booktabs}
\usepackage{subcaption}
\usepackage{graphicx}
\usepackage{comment}
\usepackage{multirow}
\usepackage{caption}
\usepackage{wrapfig}

\hypersetup{
    colorlinks=false,
    pdfborder={0 0 1},
    citebordercolor={0 1 0},
    linkbordercolor={1 0 0},
    urlbordercolor={0 0 1}
}

\title{Self-correction Optimization for Interleaved Multimodal Generation}

\iclrfinalcopy

\author{
\textbf{Xin You}$^{1}$ \quad \textbf{Zhiwen Ning}$^{1}$ \quad \textbf{Zukai Chen}$^{2}$ \quad \textbf{Minghui Zhang}$^{1}$ \quad \textbf{Xuanke Shi}$^{2}$ \\ \textbf{Hanxiao Zhang}$^{1}$  \textbf{Jingsong Liu}$^{3}$ \quad \textbf{Jie Yang}$^{1}$ \quad \textbf{Quan Wang}$^{2}$ \quad \textbf{Yun Gu}$^{1}$ \\
$^{1}$Shanghai Jiao Tong University \quad
$^{2}$SenseTime Research \quad
$^{3}$Technical University of Munich \\
\texttt{\{sjtu\_youxin, geron762\}@sjtu.edu.cn} \quad
\texttt{wangquan@sensetime.com}
}

\begin{document}

\maketitle

\thispagestyle{empty}

\begin{abstract}
Multimodal large language models (MLLMs) have made significant progress in visual understanding and generation. However, generating interleaved image–text content remains challenging, as it requires tightly integrated multimodal understanding and generation capabilities. Although existing MLLMs provide promising solutions, most rely on additional training with augmented data, which is computationally expensive and remains limited in preserving visual subjects, temporal consistency, and physical plausibility. In this work, we propose self-correction optimization (SCO), an effective training-free method for consistent interleaved generation. SCO treats the classifier-free guidance update as a reference and performs minimal self-correction under two complementary constraints, including new-event and state-preserving constraints. Specifically, the new-event constraint promotes temporal consistency across image–text sequences, while the state-preserving constraint maintains the coherence of visual subjects throughout subsequent generation steps. Experiments on challenging interleaved multimodal generation benchmarks demonstrate significant improvements in temporal coherence and visual-subject preservation. Furthermore, SCO can be extended to video generation and improves the modeling of physically grounded processes, including robot manipulation and long-horizon handcrafting.
\end{abstract}


\section{Introduction}

Multimodal large language models (MLLMs) have recently achieved substantial progress in visual understanding and generation. This progress has enabled a broader generation paradigm in which text and images are produced in an arbitrary interleaved order, while conditioning on outputs of an evolving multimodal history. Such interleaved image--text generation supports applications including visual instruction, storytelling, design assistance, multi-step image editing, multimodal planning, and embodied reasoning~\cite{ai2026llada2,shi2026duogen,alayrac2022flamingo,zheng2023minigpt,tian2024mm,team2024chameleon,wang2024emu3,chern2024anole,liu2026breaking}. 



Despite recent advances, existing interleaved generation systems still rely on multimodal context modeling, additional training data, or architectural designs to maintain cross-modal and temporal consistency \cite{tian2024mm,ye2026loom,chen2025comm,feng2026high,xing2026wan,deng2025emerging,diao2026sensenova,han2026vision,diao2026sensenova1}. However, these mechanisms are computationally expensive. Furthermore, they do not explicitly correct the generation update when the current instruction conflicts with visual states inherited from previous rounds. In particular, a model may fail to sufficiently express the newly event, or may introduce subject drift and scene reconfiguration while following the current instruction. Such errors can accumulate over successive generation rounds, thereby compromising performance in procedural tasks.

\begin{figure}[!t]
\centerline{\includegraphics[width=1.0\linewidth]{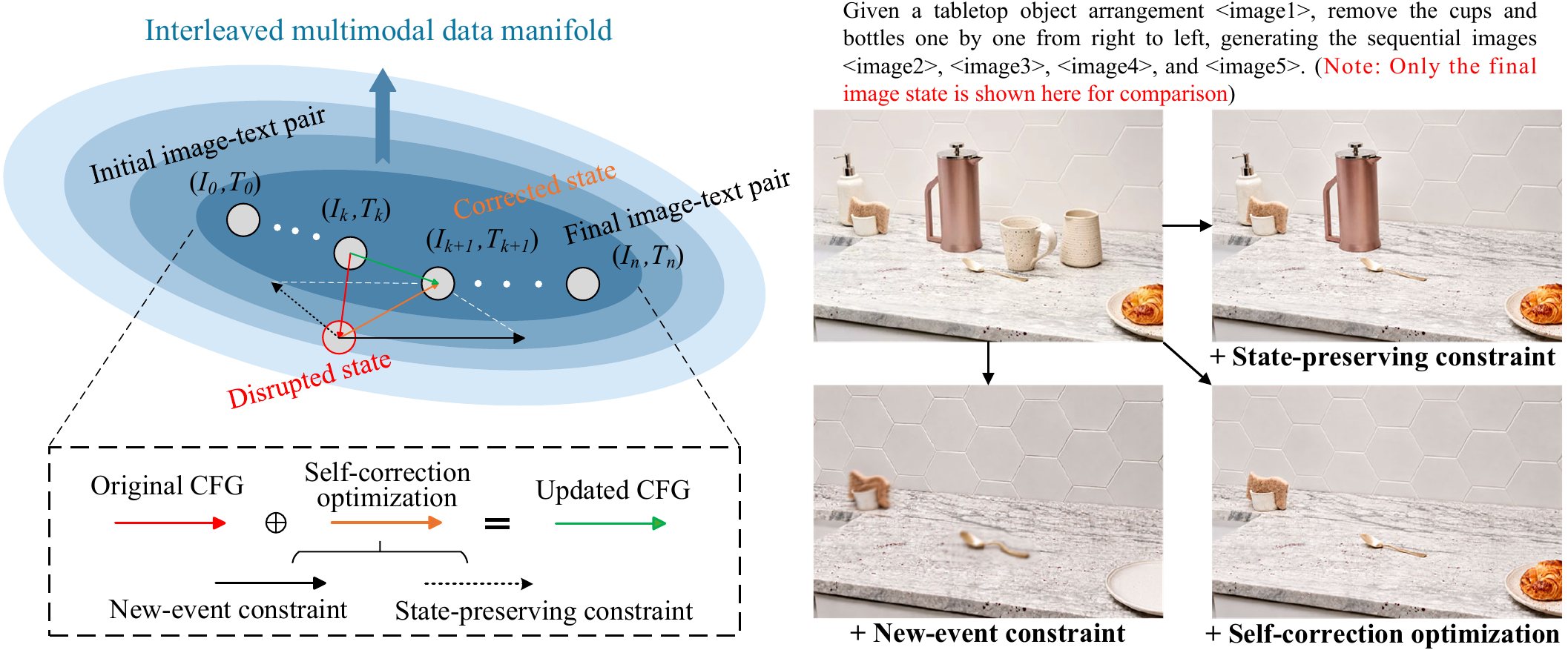}}
\caption{The mechanism of self-correction optimization, which consists of new-event and state-preserving constraints for the new-event realization and visual subject consistency.}
\label{concept_sco}
\end{figure}

In this work, we propose \emph{Self-Correction Optimization} (SCO), an effective training-free method for plausible interleaved generation. SCO augments the classifier-free guidance (CFG) with two complementary constraints. The \emph{new-event constraint} encourages the current image to realize the event specified by the current instruction and its direct visual consequences. The \emph{state-preserving constraint} encourages the image to retain persistent visual information inherited from previous generation rounds, including subject identity, scene layout, and rendering style.

As shown in Fig.\ref{concept_sco}, at each selected generation step, SCO takes the original CFG update as a reference and applies a self-correction when either constraint is insufficiently satisfied. The two constraint are constructed from complementary conditional branches and jointly optimized through a bounded two-dimensional projection. This optimization explicitly models the interaction between event realization and state preservation, while independent correction budgets provide a controlled relaxation mechanism when the two constraints compete. SCO is performed entirely at test time and requires no additional model parameters or task-specific training.

We evaluate SCO on challenging interleaved image--text generation benchmarks. Across Opening-Bench \cite{zhou2025opening} and ISG-Bench \cite{chen2025interleaved}, SCO improves temporal coherence and visual-subject preservation, demonstrating the effectiveness of test-time self-correction for interleaved multimodal generation. Besides, the same constraint formulation can be extended to video generation. The new-event constraint captures the motion, while the state-preserving constraint maintains persistent information from the preceding temporal context. Experiments on long-horizon handcrafting and robot-manipulation scenarios demonstrate that SCO consistently improves the generation quality by powerful generators. Our contributions are summarized as follows:
\begin{itemize}[leftmargin=1.2em, itemindent=0pt, labelsep=0.4em]
    \item We formulate interleaved generation as an optimization problem that jointly requires new-event realization and preservation of the evolving multimodal state.
    
    \item We propose \emph{Self-Correction Optimization} (SCO), a training-free method that jointly optimizes complementary new-event and state-preserving constraints through a bounded projection of the original CFG update, with controlled relaxation under constraint conflicts.
    
    \item Experiments on interleaved image--text and physically grounded video-generation tasks demonstrate improved event consistency, visual-subject preservation, and temporal coherence without additional training.
\end{itemize}

\section{Related Work}

\subsection{Modular Interleaved Generation Models}

Early generation-oriented approaches adopted modular architectures that couple an autoregressive language model with a pretrained visual generator. GILL~\cite{koh2023generating} connects a language model to a frozen visual generation and retrieval module through learned projections. MiniGPT-5~\cite{zheng2023minigpt} introduced generative vokens as an interface between the language model and Stable Diffusion~\cite{rombach2022high}, enabling the generation of interleaved text--image responses. MM-Interleaved~\cite{tian2024mm} further improved multimodal context modeling through multi-scale and multi-image feature synchronization, allowing the generator to access visual information from previously generated images. OpenLEAF~\cite{an2023openleaf} explored modular interleaved generation in open-domain applications, including visual instructions, multimodal storytelling, and webpage or poster creation. These modular systems provide a practical way to combine language reasoning with high-quality image synthesis and avoid training a fully unified generator. However, their inherent architectures restrict them to interleaved generation.

\subsection{Unified Multimodal Models for Interleaved Generation}

Unified multimodal models (UMMs) provide a promising paradigm by modeling language and visual content within a shared architecture. Chameleon~\cite{team2024chameleon} adopts early fusion of discretized image and text tokens and supports mixed-modal understanding and generation. Emu3~\cite{wang2024emu3} formulates language, image, and video generation under a unified next-token prediction objective, while Anole~\cite{chern2024anole} improves the accessibility of native interleaved generation through parameter- and data-efficient adaptation. DuoGen~\cite{shi2026duogen} narrows the gap between modular and unified generation by jointly adapting a vision-language model and a pretrained video generator for interleaved multimodal generation. However, these unified models may over-rely on the most recent visual state and terminate at an intermediate state that does not fully satisfy the current instruction. Small deviations introduced in early generation rounds can also accumulate across subsequent steps, leading to subject drift and incorrect visual states. That is why SCO is proposed to explicitly optimize complementary new-event and state-preserving constraints without additional task-specific training.

\subsection{Training-Free Generation Guidance}

Classifier-free guidance (CFG)~\cite{ho2022classifier} improves conditional generation by combining conditional and unconditional predictions during sampling. It is widely used in image and video diffusion models, including latent diffusion~\cite{rombach2022high}, Video Diffusion Models~\cite{ho2022video}, and Imagen Video~\cite{ho2022imagen}. However, its single scalar guidance scale offers limited control when multiple semantic objectives must be satisfied simultaneously. Several training-free methods extend guidance to multiple conditions or auxiliary objectives, including composable diffusion and Universal Guidance~\cite{bansal2024universal}. These methods improve controllability while keeping the pretrained generator fixed, providing a lightweight alternative to task-specific post-training~\cite{ye2024tfg}. The proposed SCO follows this test-time paradigm but targets cross-modal consistency and temporal coherence in interleaved generation.

\section{Methodology}

\subsection{Problem Formulation and Motivation}

We consider interleaved image--text generation, where text and image outputs
may appear in an arbitrary order and each image is conditioned on the complete
multimodal history. Let
$\mathcal{O}=(o_1,\ldots,o_L)$ denote an interleaved output with
$o_\ell\in\mathcal{T}\cup\mathcal{X}$, where $\mathcal{T}$ and $\mathcal{X}$
denote the text and image spaces, respectively. At the $m$-th image slot,
$\mathcal{H}_m$ denotes the preceding multimodal history and $e_m$ denotes
the current event or instruction. The history is updated after generating the
current image:
\begin{equation}
\begin{aligned}
\hat{x}_m
&\sim p_\theta(\,\cdot\mid\mathcal{H}_m,e_m),
&
\mathcal{H}_{m+1}
&=
\mathcal{H}_m\oplus(e_m,\hat{x}_m),\\
v_{\mathrm{ref}}(z_\tau)
&=
v_{\mathrm{u}}(z_\tau)
+
\omega
\left[
v_{\mathrm{f}}(z_\tau\mid\mathcal{H}_m,e_m)
-
v_{\mathrm{u}}(z_\tau)
\right],
&
\delta_{\mathrm{ref}}
&=
v_{\mathrm{ref}}-v_{\mathrm{u}} .
\end{aligned}
\label{eq:problem-cfg}
\end{equation}
Here, $z_\tau$ denotes the current latent state at a selected sampling or
integration step, $v_{\mathrm{u}}$ and $v_{\mathrm{f}}$ denote the
unconditional and fully conditional model outputs, and $\omega$ is the CFG
scale. The notation $v$ represents the model update, which may correspond to
a noise, score, or velocity prediction depending on the underlying generator.
We use $\tau\in[0,1]$ to denote the normalized sampling progress, with
$\tau=0$ corresponding to the beginning of the sampling trajectory.

The full condition in \eqref{eq:problem-cfg} entangles the current event with
the visual state inherited from previous rounds. Consequently, the original
CFG update does not explicitly distinguish between information that should
change and information that should remain stable. This may lead to
under-realization of the current event or progressive changes in subject
identity, scene layout, etc.

Self-Correction Optimization (SCO) addresses this issue by constructing two
complementary directions: a new-event direction that captures the
requested change and a state-preserving direction that captures persistent
visual information. SCO then minimally corrects the original CFG update when
one or both requirements are insufficiently satisfied.

\subsection{New-Event Constraint}

We first isolate the update associated with the current event. Specifically,
we define the full condition and its counterfactual counterpart as:
\begin{equation}
c_{\mathrm{full}}
=
(\mathcal{H}_m,e_m),
\qquad
c_{\mathrm{counter}}
=
(\mathcal{H}_m,e_{\varnothing}),
\label{eq:event-conditions}
\end{equation}
where $e_{\varnothing}$ describes the visual state immediately before the
current event occurs. The two conditions share the same multimodal history and
differ only in the current event specification. Evaluating them at the same
latent state yields:
\begin{equation}
v_{\mathrm{f}}
=
v_\theta(z_\tau\mid c_{\mathrm{full}}),
\qquad
v_{\mathrm{counter}}
=
v_\theta(z_\tau\mid c_{\mathrm{counter}}),
\qquad
a_{\mathrm{new}} = v_{\mathrm{f}}-v_{\mathrm{counter}} .
\label{eq:new-direction}
\end{equation}
Here $\theta$ denotes the frozen parameters of the pretrained generator, and
$v_\theta$ denotes its sampling update. The difference
$a_{\mathrm{new}}$ defines the new-event direction, which captures the
model's local response to introducing the current event while holding the
history and latent state fixed. For a candidate update $\delta$, its support for the current event is measured
by its projection onto $a_{\mathrm{new}}$:
\begin{equation}
p_{\mathrm{new}}(\delta)
=
a_{\mathrm{new}}^{\top}\delta
\geq
c_{\mathrm{new}},
\qquad
c_{\mathrm{new}}
=
\lambda_{\mathrm{new}}(\tau)
\lVert a_{\mathrm{new}}\rVert ,
\label{eq:new-constraint}
\end{equation}
where $\lambda_{\mathrm{new}}(\tau)$ controls the required event strength. When the original CFG update $\delta_{\mathrm{ref}}$ satisfies
$a_{\mathrm{new}}^{\top}\delta_{\mathrm{ref}}<c_{\mathrm{new}}$, it provides
insufficient directional support for the current event, and SCO activates the
corresponding new-event correction. This constraint reduces visual inertia and
promotes the realization of the requested event and its direct consequences.

\subsection{State-Preserving Constraint}

Realizing a new event alone is insufficient for long-horizon generation,
because the corresponding update may alter visual properties that should
remain stable. Therefore, we try to isolate the state-preserving direction
$a_{\mathrm{state}}$, by constructing an event-only condition that retains the
current event while removing the preceding multimodal history:
\begin{equation}
c_{\mathrm{event}}
=
(\varnothing,e_m),
\qquad
v_{\mathrm{event}}
=
v_\theta(z_\tau\mid c_{\mathrm{event}}),
\qquad
a_{\mathrm{state}}
=
v_{\mathrm{f}}-v_{\mathrm{event}} .
\label{eq:state-direction}
\end{equation}

The difference in \eqref{eq:state-direction} estimates the contribution of
the inherited state that is present in the full condition but absent from the
event-only condition. This state includes subject identity,
scene layout, and rendering style. The state-preserving constraint is as follows:
\begin{equation}
p_{\mathrm{state}}(\delta)
=
a_{\mathrm{state}}^{\top}\delta
\geq
c_{\mathrm{state}},
\qquad
c_{\mathrm{state}}
=
\lambda_{\mathrm{state}}(\tau)
\lVert a_{\mathrm{state}}\rVert ,
\label{eq:state-constraint}
\end{equation}
where $\lambda_{\mathrm{state}}(\tau)$ controls the state-preserving strength. Unlike direct image copying, this constraint acts in the generator-update
space and preserves semantic and visual attributes without requiring
pixel-level correspondence. It consequently reduces subject drift, layout changes, and style inconsistency across successive image slots.

\subsection{Joint Self-Correction Optimization}

The constraint strengths are scheduled over the normalized sampling trajectory.
Let $\tau_{\mathrm{end}}\in[0,1]$ denote the normalized sampling endpoint. SCO is applied from
$\tau=0$ to $\tau=\tau_{\mathrm{end}}$:
\begin{equation}
\lambda_{\mathrm{new}}(\tau)
=
\alpha_0\,
\mathbf{1}
\!\left\{
0\leq\tau\leq\tau_{\mathrm{end}}
\right\},
\qquad
\lambda_{\mathrm{state}}(\tau)
=
s_{\mathrm{state}}
\lambda_{\mathrm{new}}(\tau),
\label{eq:strength-scheduling}
\end{equation}
where $\alpha_0$ controls the overall strength of the new-event constraint and
$s_{\mathrm{state}}$ controls the relative strength of the state-preserving constraint. The two directions generally lie in the same update space and are not
orthogonal. Independently applying two corrections may therefore cause one
correction to weaken the other. SCO jointly modifies the original
CFG update $\delta_{\mathrm{ref}}$ as the refined update $\delta^*$:
\begin{equation}
\delta^*
=
\delta_{\mathrm{ref}}
+
\mu_{\mathrm{new}}a_{\mathrm{new}}
+
\mu_{\mathrm{state}}a_{\mathrm{state}},
\qquad
v^*
=
v_{\mathrm{u}}+\delta^* ,
\label{eq:corrected-update}
\end{equation}
where $\mu_{\mathrm{new}},\mu_{\mathrm{state}}\geq0$ are the correction
coefficients.

Ignoring correction budgets, the optimal update is obtained by projecting the
original CFG update onto the intersection of the two constraint half-spaces:
\begin{equation}
\begin{aligned}
\min_{\delta}\quad
&
\frac{1}{2}
\left\|
\delta-\delta_{\mathrm{ref}}
\right\|^2
\\
\mathrm{s.t.}\quad
&
a_i^{\top}\delta\geq c_i,
\qquad
i\in\{\mathrm{new},\mathrm{state}\}.
\end{aligned}
\label{eq:joint-projection}
\end{equation}

The interaction between the constraints is characterized by the Gram matrix:
\begin{equation}
G_{ij}
=
a_i^{\top}a_j,
\qquad
i,j\in\{\mathrm{new},\mathrm{state}\},
\label{eq:gram-matrix}
\end{equation}
whose off-diagonal term satisfies:
\begin{equation}
G_{\mathrm{new},\mathrm{state}}
=
\lVert a_{\mathrm{new}}\rVert
\lVert a_{\mathrm{state}}\rVert
\cos\theta,
\qquad
\cos\theta
=
\frac{
a_{\mathrm{new}}^{\top}a_{\mathrm{state}}
}{
\lVert a_{\mathrm{new}}\rVert
\lVert a_{\mathrm{state}}\rVert+\epsilon
}.
\label{eq:direction-interaction}
\end{equation}
A positive cosine indicates local cooperation between the two constraints,
whereas a negative cosine indicates competition. The off-diagonal term in
\eqref{eq:direction-interaction} is therefore essential: a correction along one
direction can change the projection onto the other. This cross-effect is
explicitly accounted for by the joint projection in
\eqref{eq:joint-projection}, whereas sequentially applying two independent
corrections may undo a constraint that has already been satisfied.

To prevent excessive deviations from the original CFG update, SCO imposes
independent correction budgets:
\begin{equation}
0\leq\mu_{\mathrm{new}}\leq M_{\mathrm{new}},
\qquad
0\leq\mu_{\mathrm{state}}\leq M_{\mathrm{state}},
\label{eq:correction-budgets}
\end{equation}
where $M_{\mathrm{new}}$ and $M_{\mathrm{state}}$ respectively bound the
correction strengths for the new-event and state-preserving directions. The resulting problem is a convex optimization over
only two correction coefficients. Therefore, its feasible active sets can be
evaluated directly, avoiding the overhead of a generic high-dimensional
quadratic-programming solver. The bounded correction also admits the following soft-constraint
interpretation:
\begin{equation}
\begin{aligned}
\min_{\delta,\xi_{\mathrm{new}},\xi_{\mathrm{state}}\geq0}
\quad&
\frac{1}{2}
\left\|
\delta-\delta_{\mathrm{ref}}
\right\|^2
+
M_{\mathrm{new}}\xi_{\mathrm{new}}
+
M_{\mathrm{state}}\xi_{\mathrm{state}}
\\
\mathrm{s.t.}\quad&
a_i^{\top}\delta+\xi_i\geq c_i,
\qquad
i\in\{\mathrm{new},\mathrm{state}\}.
\end{aligned}
\label{eq:soft-constraints}
\end{equation}
The residual relaxation of constraint $i$ is measured by
$s_i=[c_i-a_i^{\top}\delta^*]_+$. A positive $s_i$ indicates that the
constraint is not fully enforced because its correction budget is insufficient
or because it competes with the other constraint. Thus, a negative direction
cosine alone does not imply infeasibility; actual relaxation is determined by
the post-correction residual.

At each selected sampling or integration step, SCO evaluates the unconditional,
full, counterfactual, and event-only branches at the same latent state. It
then constructs the two contrastive directions, computes the scheduled
thresholds using \eqref{eq:strength-scheduling}, and solves the bounded
two-dimensional optimization defined by
\eqref{eq:joint-projection} and \eqref{eq:correction-budgets}. The resulting
update is substituted for the original CFG update according to
\eqref{eq:corrected-update}. All model parameters remain frozen, and the
entire procedure is performed at test time.

Beyond interleaved image--text generation, SCO can be extended to video
generation by treating each temporal position as an image slot. In our video
experiments, a GPT-5-based planning stage first decomposes the text prompt into
temporally ordered events and allocates the available frames to these events. The two SCO directions and the bounded
joint correction in
\eqref{eq:joint-projection}--\eqref{eq:correction-budgets} are then applied to
each frame-indexed component, including when multiple frames are generated
jointly by a video generator. Further
formulation details are provided in Appendix~\ref{app:video-extension}.


\begin{figure}[!t]
\centerline{\includegraphics[width=1.0\linewidth]{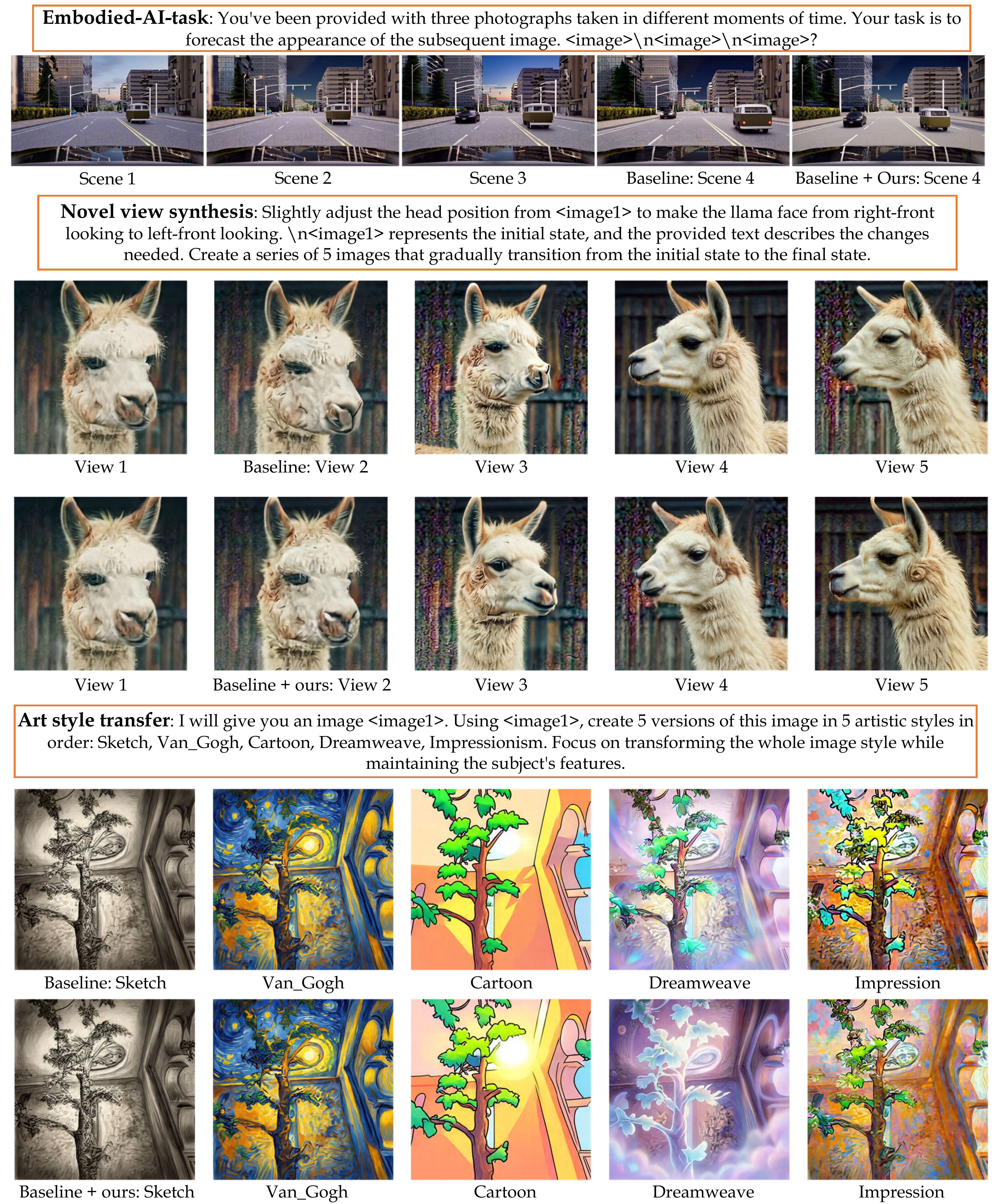}}
\caption{Qualitative results on various interleaved generation tasks when evaluating the efficacy of SCO, including iterative image editing, novel view synthesis, and embodied-AI scenarios.}
\label{qualitative_results_ablation}
\end{figure}

\section{Experiments}
\subsection{Datasets}
\textbf{Interleaved image--text generation:} We evaluate interleaved image--text generation on OpenING~\cite{zhou2025opening}
and ISG-Bench~\cite{chen2025interleaved}. OpenING contains 5,400
human-annotated instances spanning 23 real-world meta-topics and 56
fine-grained tasks, including visual planning, design, education,
travel, and creative content generation. ISG-Bench contains 1,150 samples
covering 8 scenarios and 21 subtasks. Each sample is represented by an
interleaved scene graph that specifies entities, attributes, and relations. \textbf{Video generation and embodied scenarios:}
We further evaluate SCO in video generation and embodied settings. For
long-horizon handcrafting, we use the handcraft-making subset of
Video-CraftBench introduced by VideoWorld2~\cite{ren2026videoworld}, which
contains first-person videos of multi-step tasks such as paper folding and
block building. SCO is applied on top of VideoWorld2. For robot-manipulation
videos, we use a hand--object interaction subset of the Interleaved
X-Embodiment data released with Interleave-VLA~\cite{fan2026interleave}, and
apply SCO to Cosmos-2.5 and Cosmos-3 \cite{agarwal2026cosmos}.

\subsection{Implementation details and evaluation metrics}
All selected baselines are configured according to corresponding code repositories. Experiments are implemented based on PyTorch and 2 NVIDIA H100 Tensor Core GPUs. Besides, the evaluation metrics cover four axes.
\textbf{(i) Image quality} uses FID~\cite{heusel2017gan} to measure the
distributional gap between generated and real images, and CLIP-I
(cosine similarity of CLIP~\cite{radford2021clip} image embeddings
between prediction and ground truth) for frame-level perceptual
similarity.
\textbf{(ii) Cross-modality consistency} uses CLIP-T (cosine similarity
between CLIP text and image embeddings), together with GPT-Judge 1
(GPT-J1), a GPT-5.6-sol multimodal judge that scores each frame on
prompt fidelity and factual plausibility.
\textbf{(iii) Temporal coherence} uses subject consistency $C_s$ via
DINOv2~\cite{oquab2024dinov2} features on the subject region, and
background consistency $C_b$ via CLIP features on the background.
\textbf{(iv) Fact consistency and physical plausibility} is assessed
by GPT-Judge 2 (GPT-J2), which rates whether the depicted events are
factually consistent and physically plausible.

\begin{table}[t]
\centering
\small
\caption{Quantitative results on the ISG-Bench dataset when plugging SCO into various open-source models, including modular and unified models.}
\label{tab:comparison}
\resizebox{0.96\textwidth}{!}{
\begin{tabular}{llccccccc}
\toprule
Type
& Method
& FID $\downarrow$
& CLIP-I $\uparrow$
& CLIP-T $\uparrow$
& GPT-J1 $\uparrow$
& $C_s$ $\uparrow$
& $C_b$ $\uparrow$
& GPT-J2 $\uparrow$ \\
\midrule

\multirow{6}{*}{
  \rotatebox[origin=c]{90}{\shortstack{Modular\\Model}}
}
& MiniGPT-5 \cite{zheng2023minigpt}
& 91.3 & 0.151 & 0.269 & 0.327 & 0.13 & 0.20 & 0.243 \\
& MiniGPT-5 + SCO
& 88.5 & 0.170 & 0.285 & 0.365 & 0.16 & 0.22 & 0.279 \\
\cmidrule(lr){2-9}

& Show-o2 \cite{xie2026show}
& 83.6 & 0.146 & 0.270 & 0.337 & 0.12 & 0.17 & 0.248 \\
& Show-o2 + SCO
& 79.7 & 0.169 & 0.281 & 0.349 & 0.13 & 0.19 & 0.264 \\
\cmidrule(lr){2-9}

& MM-Interleaved \cite{tian2024mm}
& 76.2 & 0.201 & 0.329 & 0.397 & 0.18 & 0.22 & 0.351 \\
& MM-Interleaved + SCO
& 73.6 & 0.227 & 0.337 & 0.419 & 0.20 & 0.23 & 0.384 \\

\midrule

\multirow{8}{*}{
  \rotatebox[origin=c]{90}{\shortstack{Unified\\Model}}
}
& BAGEL \cite{deng2025emerging}
& 83.2 & 0.195 & 0.257 & 0.306 & 0.14 & 0.19 & 0.237 \\
& BAGEL + SCO
& 78.3 & 0.211 & 0.281 & 0.329 & 0.16 & 0.20 & 0.255 \\
\cmidrule(lr){2-9}


& LLaDA2.0-Uni \cite{ai2026llada2}
& 95.8 & 0.117 & 0.226 & 0.287 & 0.09 & 0.13 & 0.189 \\
& LLaDA2.0-Uni + SCO
& 91.7 & 0.142 & 0.245 & 0.304 & 0.10 & 0.16 & 0.213 \\
\cmidrule(lr){2-9}

& SenseNova-U1 \cite{diao2026sensenova}
& 52.1 & 0.385 & 0.497 & 0.677 & 0.36 & 0.39 & 0.591 \\
& SenseNova-U1 + SCO
& 49.7 & 0.409 & 0.513 & 0.695 & 0.38 & 0.41 & 0.624 \\
\cmidrule(lr){2-9}

& DuoGen \cite{shi2026duogen}
& 45.7 & 0.413 & 0.531 & 0.692 & 0.39 & 0.43 & 0.621 \\
& DuoGen + SCO
& 41.4 & 0.445 & 0.560 & 0.735 & 0.41 & 0.47 & 0.659 \\
\bottomrule
\end{tabular}}
\end{table}

\begin{table*}[t]
\centering
\small
\caption{Quantitative results on the Opening dataset when plugging SCO into various open-source models, including modular and unified models.}
\label{tab:ipd_results}
\resizebox{0.96\textwidth}{!}{
\begin{tabular}{llccccccc}
\toprule
Type
& Method
& FID $\downarrow$
& CLIP-I $\uparrow$
& CLIP-T $\uparrow$
& GPT-J1 $\uparrow$
& $C_s$ $\uparrow$
& $C_b$ $\uparrow$
& GPT-J2 $\uparrow$ \\
\midrule

\multirow{6}{*}{
  \rotatebox[origin=c]{90}{\shortstack{Modular\\Model}}
}
& MiniGPT-5 \cite{zheng2023minigpt}
& 90.4 & 0.149 & 0.257 & 0.445 & 0.18 & 0.23 & 0.344 \\
& MiniGPT-5 + SCO
& 89.9 & 0.138 & 0.247 & 0.423 & 0.17 & 0.21 & 0.319 \\
\cmidrule(lr){2-9}

& Show-o2 \cite{xie2026show}
& 88.5 & 0.173 & 0.280 & 0.471 & 0.19 & 0.26 & 0.385 \\
& Show-o2 + SCO
& 84.0 & 0.179 & 0.291 & 0.496 & 0.19 & 0.28 & 0.401 \\
\cmidrule(lr){2-9}

& MM-Interleaved \cite{tian2024mm}
& 94.1 & 0.140 & 0.269 & 0.432 & 0.15 & 0.19 & 0.327 \\
& MM-Interleaved + SCO
& 81.3 & 0.164 & 0.293 & 0.450 & 0.17 & 0.22 & 0.369 \\

\midrule

\multirow{8}{*}{
  \rotatebox[origin=c]{90}{\shortstack{Unified\\Model}}
}
& BAGEL \cite{deng2025emerging}
& 87.2 & 0.190 & 0.303 & 0.459 & 0.21 & 0.23 & 0.374 \\
& BAGEL + SCO
& 83.7 & 0.217 & 0.339 & 0.482 & 0.23 & 0.26 & 0.398 \\
\cmidrule(lr){2-9}

& LLaDA2.0-Uni \cite{ai2026llada2}
& 118.6 & 0.083 & 0.219 & 0.237 & 0.05 & 0.11 & 0.185 \\
& LLaDA2.0-Uni + SCO
& 108.4 & 0.089 & 0.225 & 0.234 & 0.07 & 0.10 & 0.249 \\
\cmidrule(lr){2-9}

& SenseNova-U1 \cite{diao2026sensenova}
& 67.2 & 0.325 & 0.471 & 0.658 & 0.29 & 0.37 & 0.574 \\
& SenseNova-U1 + SCO
& 55.9
& 0.376
&0.507
& 0.694
&0.35
& 0.40
& 0.602 \\
\cmidrule(lr){2-9}

& DuoGen \cite{shi2026duogen}
& 53.6 & 0.382 & 0.505 & 0.683 & 0.34 & 0.40 & 0.603 \\
& DuoGen + SCO
& 50.1 & 0.405 & 0.539 & 0.701 & 0.36 & 0.42 & 0.637 \\

\bottomrule
\end{tabular}}
\end{table*}

\subsection{Experimental Results}

Tables~\ref{tab:comparison} and~\ref{tab:ipd_results} summarize the
performance of SCO on OpenING and ISG-Bench, respectively. Across both
benchmarks, SCO generally improves visual quality, cross-modal alignment, and
cross-image consistency for both modular and unified models. On ISG-Bench, the
concurrent improvements in $C_s$, $C_b$, and GPT-J2 indicate that SCO enhances
current-event realization while preserving the visual state established in
previous image slots. DuoGen~\cite{shi2026duogen} achieves the largest relative
FID reduction ($9.4\%$) and the strongest overall performance after applying
SCO. On OpenING, the most pronounced image-quality improvements are observed
for SenseNova-U1~\cite{diao2026sensenova} and MM-Interleaved~\cite{tian2024mm},
whereas DuoGen with SCO provides the strongest overall balance between
cross-modal alignment and sequence-level consistency. SCO also improves the
holistic GPT-J2 score of LLaDA2.0-Uni~\cite{ai2026llada2}, despite variations
in individual metrics. MiniGPT-5~\cite{zheng2023minigpt} is an exception,
showing a slight FID improvement but degradation on the remaining metrics.
This result highlights that the effectiveness of SCO depends on the quality
and separability of the full, counterfactual, and event-only conditional
branches. Moreover, we present qualitative results by plugging SCO into Sensenova-U1. As shown in Fig.\ref{qualitative_results_ablation}, SCO brings significant improvements to various tasks including iterative image-editing, novel view synthesis, embodied AI, etc. Overall, SCO serves as a general training-free correction layer that
improves interleaved generation across different model architectures without
additional task-specific training.

\begin{figure}[!t]
\centerline{\includegraphics[width=1.0\linewidth]{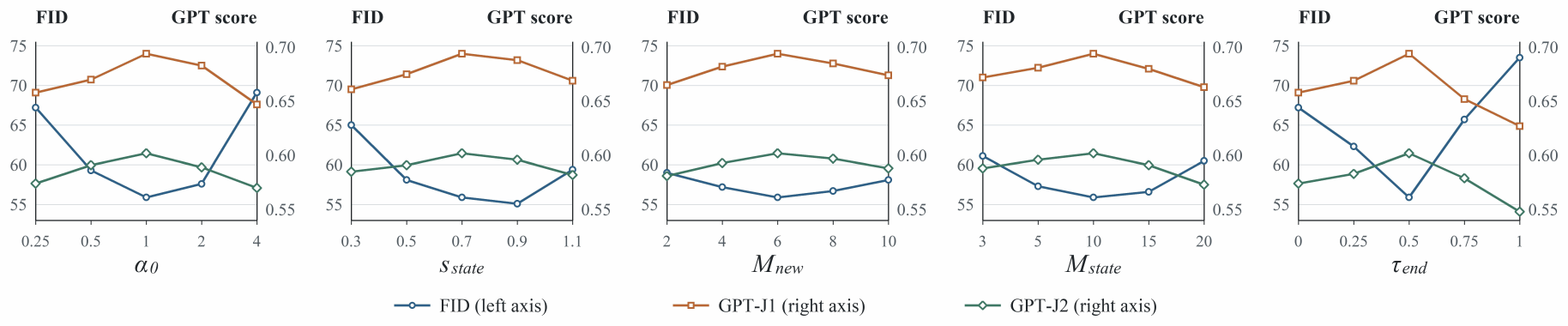}}
\caption{Sensitivity analysis for hyper-parameters of SCO ($\alpha_{0}$, $s_{state}$, $M_{state}$, $M_{new}$, and $\tau_{end}$).}
\label{hyper analysis}
\end{figure}

\subsection{Comparison with CFG-based methods}
\label{sec:cfg_comparison}

\noindent
\begin{minipage}[t]{0.37\columnwidth}
\vspace{0pt}
\small
To assess the effectiveness of the proposed SCO, we conduct
experiments on the Opening dataset, using SenseNova-U1 as the
baseline. SCO achieves the best performance across all
four metrics. As shown in Table \ref{tab:guidance_comparison}, compared with the baseline, SCO reduces FID by
16.8\% and improves CLIP-I, CLIP-T, and GPT-J2 by 0.051, 0.036,
and 0.028, respectively.
\end{minipage}
\hfill
\begin{minipage}[t]{0.60\columnwidth}
\vspace{0pt}
\centering
\captionsetup{font=small,skip=4pt}
\captionof{table}{Quantitative comparison of different CFG-based methods.
Best results are in bold.}
\label{tab:guidance_comparison}

\resizebox{\linewidth}{!}{%
\begin{tabular}{lcccc}
\toprule
Method
& FID $\downarrow$
& CLIP-I $\uparrow$
& CLIP-T $\uparrow$
& GPT-J2 $\uparrow$ \\
\midrule
Baseline
& 67.2 & 0.325 & 0.471 & 0.574 \\
+ Interleaved CFG \cite{wang2026illuminating}
& 62.7 & 0.343 & 0.485 & 0.589 \\
+ CFG++ \cite{chung2025cfg++}
& 60.4 & 0.358 & 0.481 & 0.577 \\
+ TCFG \cite{fu2025tcfg}
& 59.2 & 0.349 & 0.484 & 0.583 \\
+ SCO (Ours)
& \textbf{55.9}
& \textbf{0.376}
& \textbf{0.507}
& \textbf{0.602} \\
\bottomrule
\end{tabular}%
}
\end{minipage}

Compared with CFG++ \cite{chung2025cfg++}, SCO further reduces FID from 60.4 to 55.9,
corresponding to a 7.5\% relative reduction, while improving
CLIP-I, CLIP-T, and GPT-J2 by 0.018, 0.026, and 0.025,
respectively. These consistent gains suggest that SCO's ability
to introduce new events while preserving historical states
contributes to improved visual fidelity and text-image alignment.

\subsection{Hyper-parameter Analysis}
We use SenseNova-U1 as the baseline and conduct hyperparameter analysis on the OpenING dataset. The quantitative results in Fig.\ref{hyper analysis} show that moderate constraint settings provide the best balance between preserving existing states and allowing valid changes. For the new-event constraint, $\alpha_0=1.0$ and $M_{\mathrm{new}}=6$ achieve the strongest overall performance. For the state constraint, the optimal region for $s_{\mathrm{state}}$ approximately belongs to $[0.7\mbox{-}0.9]$, and the optimal $M_{\mathrm{state}}$ varies around $10$. Weak constraints can cause identity, layout, background, or style drift, whereas overly strong constraints may suppress legitimate state changes.

\begin{figure}[t]
\centering

\begin{minipage}[t]{0.48\linewidth}
\vspace{0pt}
\centering

\includegraphics[width=\linewidth]{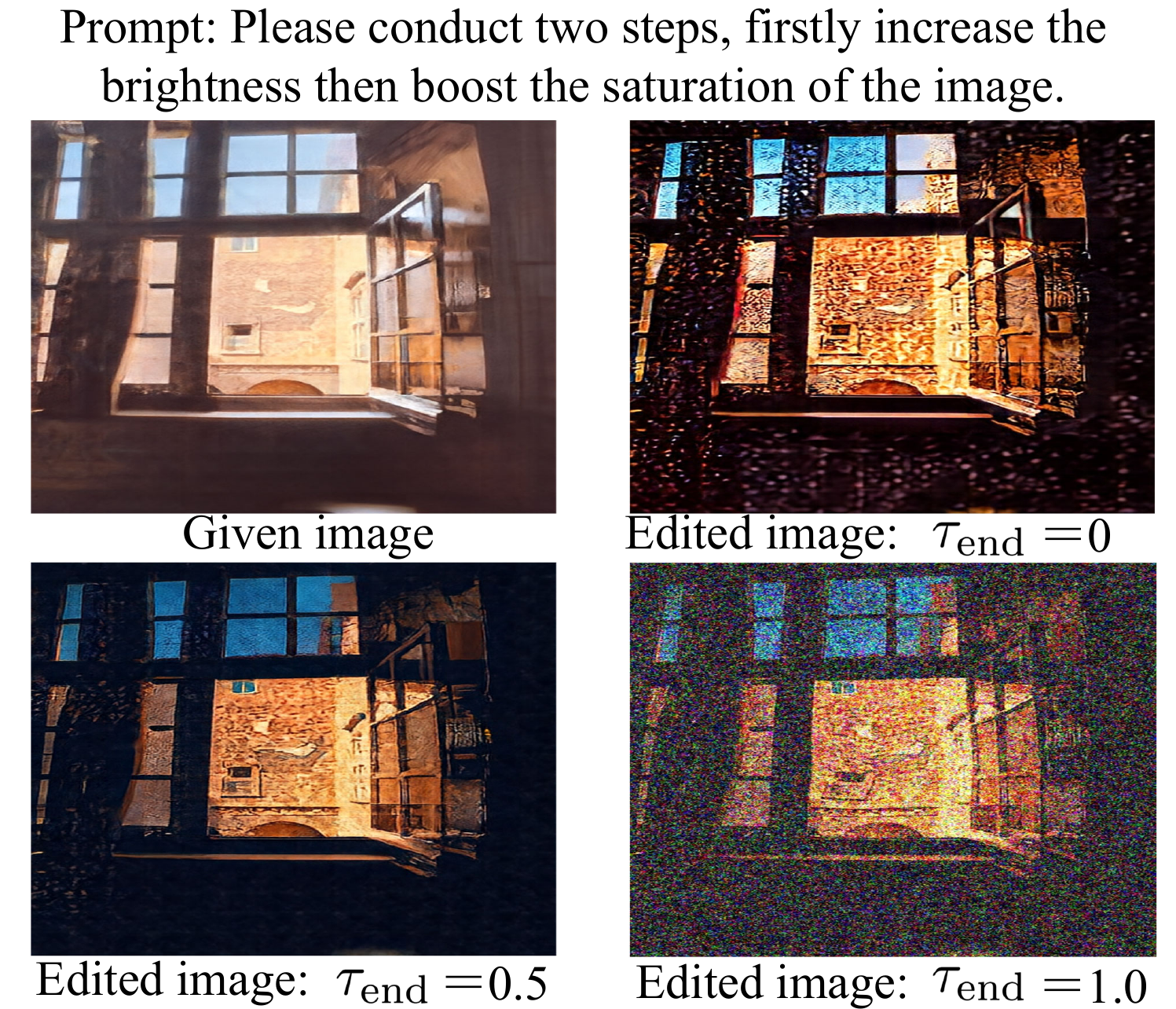}

\captionof{figure}{The choice for
$\tau_{\mathrm{end}}$.}
\label{fig:tau_end}
\end{minipage}
\hfill
\begin{minipage}[t]{0.49\linewidth}
\vspace{0pt}
\centering
\scriptsize
\setlength{\tabcolsep}{2pt}

\captionof{table}{Video generation performance evaluation for the efficacy of SCO.
VideoWorld2~\cite{ren2026videoworld} and Cosmos-2.5/Cosmos-3
~\cite{ali2025world,agarwal2026cosmos} are used as baselines for the
handcrafting and Interleaved-VLA datasets respectively.}
\label{tab:video_evaluation}

\resizebox{\linewidth}{!}{%
\begin{tabular}{lccccc}
\toprule
Model
& FVD $\downarrow$
& LPIPS $\downarrow$
& $C_s$ $\uparrow$
& $C_b$ $\uparrow$
& GPT-J2 $\uparrow$ \\
\midrule

VideoWorld2
& 178.0 & 2.487 & 0.26 & 0.38 & 0.527 \\
\textit{\quad + SCO}
& 146.3 & 1.708 & 0.30 & 0.41 & 0.562 \\

\midrule

Cosmos-2.5
& 167.2 & 1.729 & 0.37 & 0.48 & 0.583 \\
\textit{\quad + SCO}
& 142.3 & 1.337 & 0.41 & 0.52 & 0.640 \\

Cosmos-3
& 138.9 & 1.174 & 0.49 & 0.56 & 0.651 \\
\textit{\quad + SCO}
& 119.3 & 1.026 & 0.53 & 0.59 & 0.705 \\

\bottomrule
\end{tabular}%
}
\end{minipage}

\end{figure}

The optimal termination time is $\tau_{\mathrm{end}}=0.5$. When $\tau_{\mathrm{end}}=0$, SCO constraints are not applied and the model degenerates to the SenseNova-U1 baseline. As shown by qualitative results in Fig.\ref{fig:tau_end}, increasing $\tau_{\mathrm{end}}$ too far improves constraint coverage but can impair image details and valid temporal changes, while a small value of $\tau_{\mathrm{end}}$ also struggles to function well. Overall, the results support using moderate constraint strengths, upper bounds, and termination times.

\begin{table*}[t]
\centering
\small
\caption{Ablation analysis of two key constraints. MM-interleaved and SenseNova-U1 are respectively adopted as the baseline model on the Opening and ISG-Bench datasets.}
\label{tab:ablation_combined}
\resizebox{0.96\textwidth}{!}{
\begin{tabular}{lccccc|ccccc}
\toprule
&
\multicolumn{5}{c|}{MM-Interleaved}
&
\multicolumn{5}{c}{SenseNova-U1}
\\
\cmidrule(lr){2-6}
\cmidrule(lr){7-11}
Setting
& FID $\downarrow$
& CLIP-I $\uparrow$
& GPT-J1 $\uparrow$
& $C_{\mathrm{s}}$ $\uparrow$
& GPT-J2 $\uparrow$
& FID $\downarrow$
& CLIP-I $\uparrow$
& GPT-J1 $\uparrow$
& $C_{\mathrm{s}}$ $\uparrow$
& GPT-J2 $\uparrow$
\\
\midrule

Baseline
& 94.1
& 0.140
& 0.432
& 0.15
& 0.327
& 52.1
& 0.385
& 0.677
& 0.36
& 0.591
\\

$+a_{\mathrm{state}}$
& 85.6
& 0.149
& 0.445
& \textbf{0.18}
& 0.320
& 51.1
& 0.393
& 0.689
& \textbf{0.39}
& 0.620
\\

$+a_{\mathrm{new}}$
& 89.4
& 0.153
& 0.440
& 0.13
& 0.362
& 50.6
& 0.397
& 0.686
& 0.34
& 0.563
\\

$+\mathrm{Both}$
& \textbf{81.3}
& \textbf{0.164}
& \textbf{0.450}
& 0.17
& \textbf{0.369}
& \textbf{49.7}
& \textbf{0.409}
& \textbf{0.695}
& 0.38
& \textbf{0.624}
\\

\bottomrule
\end{tabular}}
\end{table*}

\begin{figure}[!b]
\centerline{\includegraphics[width=0.95\linewidth]{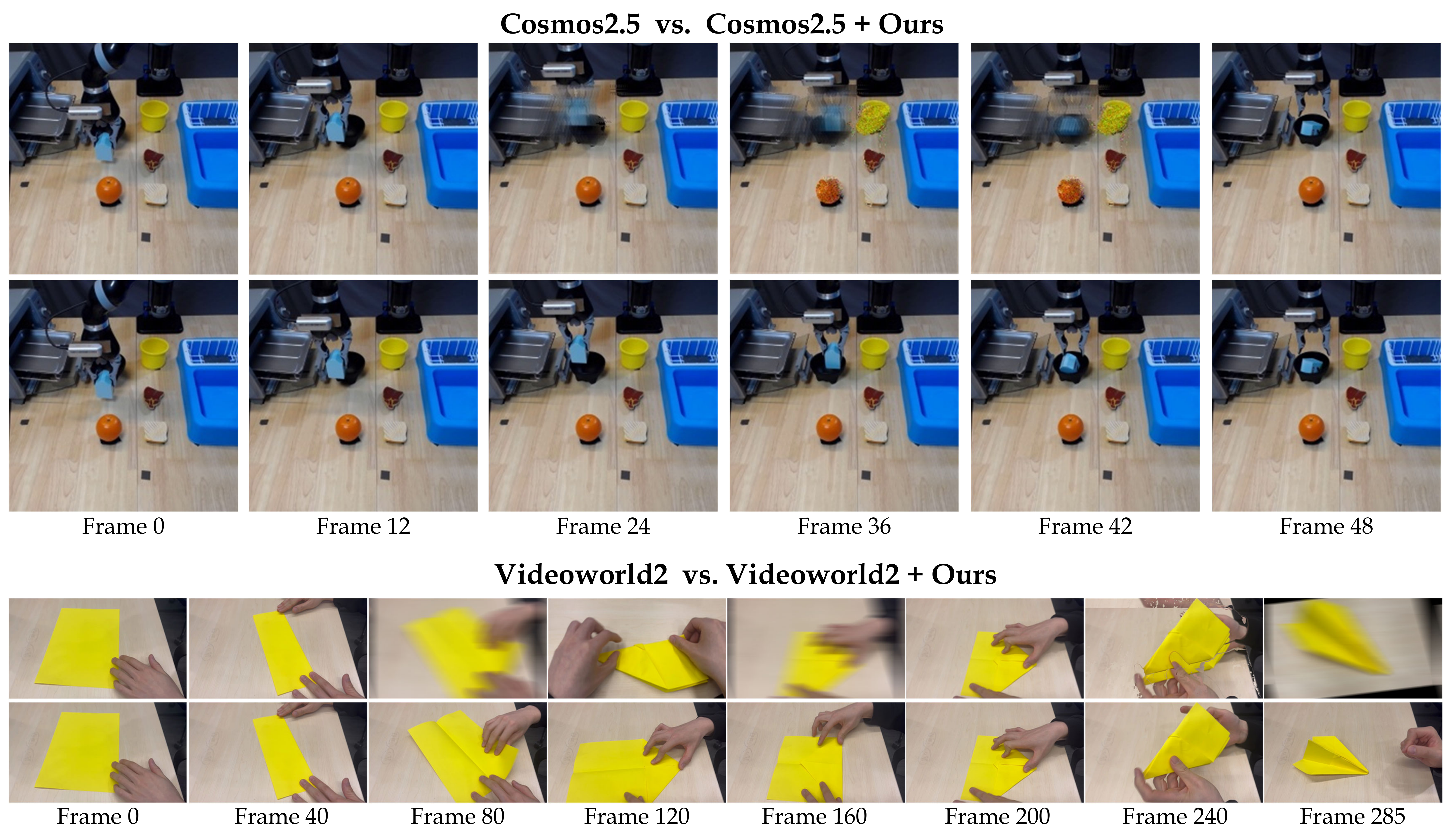}}
\caption{Qualitative results when applied to robotic arm manipulation and handcrafting scenarios.}
\label{physical_experiment}
\end{figure}

\subsection{Ablation Study on Key Components}
We implement an ablation study on two complementary constraints of SCO. Specifically, evaluations are conducted on the Opening and ISG-Bench using MM-Interleaved and
SenseNova-U1 respectively as the underlying generators, as reported in
Table~\ref{tab:ablation_combined}. The state-preserving constraint consistently
improves $C_s$, confirming its effectiveness in maintaining subject
consistency across generated images. The new-event constraint primarily
improves event realization and overall generation quality, but may introduce
greater visual variation across image slots. Jointly optimizing both
constraints achieves the best overall performance in FID, CLIP-I, GPT-J1, and
GPT-J2 for both generators. Nevertheless, its $C_s$ is slightly lower than
that of the state-preserving-only variant, since realizing newly introduced
events can inevitably alter visual content and thereby increase the difficulty
of preserving subject consistency. These results demonstrate that the two
constraints are complementary, while also revealing the inherent trade-off
between event realization and state preservation.

\subsection{Generalization on video generation}
We further evaluate SCO in two complicated video-generation scenarios: long-horizon
handcrafting and robot manipulation. Specifically, VideoWorld2 \cite{ren2026videoworld} is
evaluated on the handcrafting dataset, while Cosmos-2.5 \cite{ali2025world} and Cosmos-3 \cite{agarwal2026cosmos} are
evaluated on Interleaved-VLA. As shown in
Table~\ref{tab:video_evaluation}, integrating SCO consistently reduces FVD and
LPIPS while improving $C_s$, $C_b$, and GPT-J2 across all three generators.
The improvements in subject and background consistency indicate that the
state-preserving constraint effectively maintains persistent visual
information across temporal stages. Meanwhile, the concurrent improvement in
GPT-J2 suggests that this consistency is achieved without suppressing the
motion or physical events specified by the video instruction. Furthermore, qualitative results in Fig.\ref{physical_experiment} reveal SCO's effectiveness on preserving the state of visual subject. These results
demonstrate that SCO generalizes beyond interleaved image--text generation and
provides effective test-time control for both handcrafting and embodied
manipulation videos.

\section{Conclusion}
We introduced Self-Correction Optimization (SCO), a training-free test-time
method for improving interleaved image--text generation. SCO constructs
complementary new-event and state-preserving constraints and jointly enforces
them through a bounded minimal correction of the original CFG update.
Experiments on OpenING and ISG-Bench demonstrate broad improvements in visual
quality, cross-modal alignment, and cross-image consistency across modular and
unified generators. SCO also extends to video generation, improving temporal
consistency in handcrafting and robot-manipulation scenarios. These findings
demonstrate the importance of explicitly balancing event realization and state
preservation in long-horizon multimodal generation.

\subsection*{AI use statement}
In this work, we used generative AI tools only to edit the manuscript for readability and to assist with formatting the LaTeX source. We have not used generative AI tools to generate synthetic data, to design or execute the experiments, to implement the method, or to interpret the results. All AI-assisted edits were reviewed by the authors, and all technical claims, experimental results, figures, and tables were produced and verified by the authors.

\subsection*{Reproducibility statement}

We have present the complete theory related to SCO. Besides, the detailed process of text prompt rewriting is displayed in the appendix. All relevant hyper-parameters have been listed out in the manuscript. All above operations ensure the reproducibility of our work.

\bibliography{iclr2027_conference}

@article{alayrac2022flamingo,
  title={Flamingo: a visual language model for few-shot learning},
  author={Alayrac, Jean-Baptiste and Donahue, Jeff and Luc, Pauline and Miech, Antoine and Barr, Iain and Hasson, Yana and Lenc, Karel and Mensch, Arthur and Millican, Katherine and Reynolds, Malcolm and others},
  journal={Advances in neural information processing systems},
  volume={35},
  pages={23716--23736},
  year={2022}
}

@article{deng2025emerging,
  title={Emerging properties in unified multimodal pretraining},
  author={Deng, Chaorui and Zhu, Deyao and Li, Kunchang and Gou, Chenhui and Li, Feng and Wang, Zeyu and Zhong, Shu and Yu, Weihao and Nie, Xiaonan and Song, Ziang and others},
  journal={arXiv preprint arXiv:2505.14683},
  year={2025}
}

@article{diao2026sensenova,
  title={Sensenova-u1: Unifying multimodal understanding and generation with neo-unify architecture},
  author={Diao, Haiwen and Wu, Penghao and Deng, Hanming and Wang, Jiahao and Bai, Shihao and Wu, Silei and Fan, Weichen and Ye, Wenjie and Tong, Wenwen and Fan, Xiangyu and others},
  journal={arXiv preprint arXiv:2605.12500},
  year={2026}
}

@inproceedings{wang2026illuminating,
  title={Illuminating Unified Multimodal Model for Free-form Interleaved Text-Image Generation},
  author={Wang, Chonghuinan and Chen, Zhikai and Wang, Chunwei and Wan, Yecong and Yang, Junwei and Wang, Zhixin and Zhang, Wei and Xu, Jiaqi and Pei, Renjing and Wu, Xiaohe and others},
  booktitle={European Conference on Computer Vision},
  pages={535--552},
  year={2026},
  organization={Springer}
}

@inproceedings{chung2025cfg++,
  title={Cfg++: Manifold-constrained classifier free guidance for diffusion models},
  author={Chung, Hyungjin and Kim, Jeongsol and Park, Geon Yeong and Nam, Hyelin and Ye, Jong Chul},
  booktitle={International Conference on Learning Representations},
  volume={2025},
  pages={30824--30850},
  year={2025}
}

@inproceedings{fu2025tcfg,
  title={TCFG: Truncated Classifier-Free Guidance for Efficient and Scalable Text-to-Image Acceleration},
  author={Fu, Xiaomeng and Li, Jia},
  booktitle={2025 IEEE/CVF International Conference on Computer Vision (ICCV)},
  pages={18552--18562},
  year={2025},
  organization={IEEE}
}

@article{diao2026sensenova1,
  title={Sensenova-u1. 5: Towards native unified visual intelligence},
  author={Diao, Haiwen and Wang, Jiahao and Ding, Chenjing and Deng, Hanming and Chen, Jiangnan and Zhang, Ruixi and Wang, Ruohui and Tong, Wenwen and Fan, Xiangyu and Wang, Yubo and others},
  journal={arXiv preprint arXiv:2609.11929},
  year={2026}
}

@article{xie2026show,
  title={Show-o2: Improved native unified multimodal models},
  author={Xie, Jinheng and Yang, Zhenheng and Shou, Mike Zheng},
  journal={Advances in Neural Information Processing Systems},
  volume={38},
  pages={47490--47518},
  year={2026}
}

@article{zheng2023minigpt,
  title={Minigpt-5: Interleaved vision-and-language generation via generative vokens},
  author={Zheng, Kaizhi and He, Xuehai and Wang, Xin Eric},
  journal={arXiv preprint arXiv:2310.02239},
  year={2023}
}

@article{tian2024mm,
  title={Mm-interleaved: Interleaved image-text generative modeling via multi-modal feature synchronizer},
  author={Tian, Changyao and Zhu, Xizhou and Xiong, Yuwen and Wang, Weiyun and Chen, Zhe and Wang, Wenhai and Chen, Yuntao and Lu, Lewei and Lu, Tong and Zhou, Jie and others},
  journal={arXiv preprint arXiv:2401.10208},
  year={2024}
}

@article{an2023openleaf,
  title={Openleaf: Open-domain interleaved image-text generation and evaluation},
  author={An, Jie and Yang, Zhengyuan and Li, Linjie and Wang, Jianfeng and Lin, Kevin and Liu, Zicheng and Wang, Lijuan and Luo, Jiebo},
  journal={arXiv preprint arXiv:2310.07749},
  year={2023}
}

@inproceedings{zhou2025opening,
  title={Opening: A comprehensive benchmark for judging open-ended interleaved image-text generation},
  author={Zhou, Pengfei and Peng, Xiaopeng and Song, Jiajun and Li, Chuanhao and Xu, Zhaopan and Yang, Yue and Guo, Ziyao and Zhang, Hao and Lin, Yuqi and He, Yefei and others},
  booktitle={2025 IEEE/CVF Conference on Computer Vision and Pattern Recognition (CVPR)},
  pages={56--66},
  year={2025},
  organization={IEEE}
}

@inproceedings{chen2025interleaved,
  title={Interleaved scene graphs for interleaved text-and-image generation assessment},
  author={Chen, Dongping and Chen, Ruoxi and Pu, Shu and Liu, Zhaoyi and Wu, Yanru and Chen, Caixi and Liu, Benlin and Huang, Yue and Wan, Yao and Zhou, Pan and others},
  booktitle={International Conference on Learning Representations},
  volume={2025},
  pages={74693--74756},
  year={2025}
}

@article{xing2026wan,
  title={Wan-Weaver: Interleaved Multi-modal Generation via Decoupled Training},
  author={Xing, Jinbo and Jiang, Zeyinzi and Tuo, Yuxiang and Mao, Chaojie and Gai, Xiaotang and Chen, Xi and Zhang, Jingfeng and Pan, Yulin and Han, Zhen and Xiao, Jie and others},
  journal={arXiv preprint arXiv:2603.25706},
  year={2026}
}

@inproceedings{feng2026high,
  title={A High Quality Dataset and Reliable Evaluation for Interleaved Image-Text Generation},
  author={Feng, Yukang and Sun, Jianwen and Li, Chuanhao and Li, Zizhen and Ai, Jiaxin and Zhang, Fanrui and Zhou, Sizhuo and Chang, Yifan and Zhang, Shenglin and Dai, Yu and others},
  booktitle={International Conference on Learning Representations},
  volume={2026},
  pages={40398--40437},
  year={2026}
}

@article{team2024chameleon,
  title={Chameleon: Mixed-modal early-fusion foundation models},
  author={Team, Chameleon},
  journal={arXiv preprint arXiv:2405.09818},
  year={2024}
}

@inproceedings{chen2025comm,
  title={Comm: A coherent interleaved image-text dataset for multimodal understanding and generation},
  author={Chen, Wei and Li, Lin and Yang, Yongqi and Wen, Bin and Yang, Fan and Gao, Tingting and Wu, Yu and Chen, Long},
  booktitle={2025 IEEE/CVF Conference on Computer Vision and Pattern Recognition (CVPR)},
  pages={8073--8082},
  year={2025},
  organization={IEEE}
}

@inproceedings{ye2026loom,
  title={Loom: Diffusion-transformer for interleaved generation},
  author={Ye, Mingcheng and Liu, Jiaming and Song, Yiren},
  booktitle={Proceedings of the IEEE/CVF Conference on Computer Vision and Pattern Recognition},
  pages={4582--4592},
  year={2026}
}

@article{wang2024emu3,
  title={Emu3: Next-token prediction is all you need},
  author={Wang, Xinlong and Zhang, Xiaosong and Luo, Zhengxiong and Sun, Quan and Cui, Yufeng and Wang, Jinsheng and Zhang, Fan and Wang, Yueze and Li, Zhen and Yu, Qiying and others},
  journal={arXiv preprint arXiv:2409.18869},
  year={2024}
}

@article{chern2024anole,
  title={Anole: An open, autoregressive, native large multimodal models for interleaved image-text generation},
  author={Chern, Ethan and Su, Jiadi and Ma, Yan and Liu, Pengfei},
  journal={arXiv preprint arXiv:2407.06135},
  year={2024}
}

@article{ren2026videoworld,
  title={Videoworld 2: Learning transferable knowledge from real-world videos},
  author={Ren, Zhongwei and Wei, Yunchao and Yu, Xiao and Luo, Guixun and Zhao, Yao and Kang, Bingyi and Feng, Jiashi and Jin, Xiaojie},
  journal={arXiv preprint arXiv:2602.10102},
  year={2026}
}

@inproceedings{fan2026interleave,
  title={Interleave-vla: Enhancing robot manipulation with image-text interleaved instructions},
  author={Fan, Cunxin and Jia, Xiaosong and Sun, Yihang and Wang, Yixiao and Wei, Jianglan and Gong, Ziyang and Zhao, Xiangyu and Tomizuka, Masayoshi and Yang, Xue and Yan, Junchi and others},
  booktitle={International Conference on Learning Representations},
  volume={2026},
  pages={40872--40902},
  year={2026}
}

@inproceedings{o2024open,
  title={Open x-embodiment: Robotic learning datasets and rt-x models: Open x-embodiment collaboration 0},
  author={O’Neill, Abby and Rehman, Abdul and Maddukuri, Abhiram and Gupta, Abhishek and Padalkar, Abhishek and Lee, Abraham and Pooley, Acorn and Gupta, Agrim and Mandlekar, Ajay and Jain, Ajinkya and others},
  booktitle={2024 IEEE International Conference on Robotics and Automation (ICRA)},
  pages={6892--6903},
  year={2024},
  organization={IEEE}
}

@article{ali2025world,
  title={World simulation with video foundation models for physical ai},
  author={Ali, Arslan and Bai, Junjie and Bala, Maciej and Balaji, Yogesh and Blakeman, Aaron and Cai, Tiffany and Cao, Jiaxin and Cao, Tianshi and Cha, Elizabeth and Chao, Yu-Wei and others},
  journal={arXiv preprint arXiv:2511.00062},
  year={2025}
}

@inproceedings{zhu2025irasim,
  title={Irasim: A fine-grained world model for robot manipulation},
  author={Zhu, Fangqi and Wu, Hongtao and Guo, Song and Liu, Yuxiao and Cheang, Chilam and Kong, Tao},
  booktitle={2025 IEEE/CVF International Conference on Computer Vision (ICCV)},
  pages={9834--9844},
  year={2025},
  organization={IEEE}
}

@article{lynch2023interactive,
  title={Interactive language: Talking to robots in real time},
  author={Lynch, Corey and Wahid, Ayzaan and Tompson, Jonathan and Ding, Tianli and Betker, James and Baruch, Robert and Armstrong, Travis and Florence, Pete},
  journal={IEEE Robotics and Automation Letters},
  year={2023},
  publisher={IEEE}
}

@inproceedings{rombach2022high,
  title={High-Resolution Image Synthesis with Latent Diffusion Models},
  author={Rombach, Robin and Blattmann, Andreas and Lorenz, Dominik and Esser, Patrick and Ommer, Björn},
  booktitle={Proceedings of the IEEE/CVF Conference on Computer Vision and Pattern Recognition},
  year={2022}
}

@article{ai2026llada2,
  title={Llada2. 0-uni: Unifying multimodal understanding and generation with diffusion large language model},
  author={AI, Inclusion and Bie, Tiwei and Chen, Haoxing and Chen, Tieyuan and Cheng, Zhenglin and Cui, Long and Gan, Kai and Huang, Zhicheng and Lan, Zhenzhong and Li, Haoquan and others},
  journal={arXiv preprint arXiv:2604.20796},
  year={2026}
}

@article{shi2026duogen,
  title={Duogen: Towards general purpose interleaved multimodal generation},
  author={Shi, Min and Zeng, Xiaohui and Huang, Jiannan and Cui, Yin and Ferroni, Francesco and Li, Jialuo and Pachori, Shubham and Li, Zhaoshuo and Balaji, Yogesh and Wang, Haoxiang and others},
  journal={arXiv preprint arXiv:2602.00508},
  year={2026}
}

@inproceedings{ho2022video,
  title={Video Diffusion Models},
  author={Ho, Jonathan and Salimans, Tim and Gritsenko, Alexey
          and Chan, William and Norouzi, Mohammad and Fleet, David J.},
  booktitle={Advances in Neural Information Processing Systems},
  volume={35},
  pages={8633--8646},
  year={2022},
  url={https://arxiv.org/abs/2204.03458}
}

@article{koh2023generating,
  title={Generating images with multimodal language models},
  author={Koh, Jing Yu and Fried, Daniel and Salakhutdinov, Russ R},
  journal={Advances in Neural Information Processing Systems},
  volume={36},
  pages={21487--21506},
  year={2023}
}

@article{ho2022classifier,
  title={Classifier-free diffusion guidance},
  author={Ho, Jonathan and Salimans, Tim},
  journal={arXiv preprint arXiv:2207.12598},
  year={2022}
}

@article{ho2022imagen,
  title={Imagen video: High definition video generation with diffusion models},
  author={Ho, Jonathan and Chan, William and Saharia, Chitwan and Whang, Jay and Gao, Ruiqi and Gritsenko, Alexey and Kingma, Diederik P and Poole, Ben and Norouzi, Mohammad and Fleet, David J and others},
  journal={arXiv preprint arXiv:2210.02303},
  year={2022}
}

@inproceedings{bansal2024universal,
  title={Universal guidance for diffusion models},
  author={Bansal, Arpit and Chu, Hong-Min and Schwarzschild, Avi and Sengupta, Roni and Goldblum, Micah and Geiping, Jonas and Goldstein, Tom},
  booktitle={International Conference on Learning Representations},
  volume={2024},
  pages={51304--51323},
  year={2024}
}

@article{ye2024tfg,
  title={Tfg: Unified training-free guidance for diffusion models},
  author={Ye, Haotian and Lin, Haowei and Han, Jiaqi and Xu, Minkai and Liu, Sheng and Liang, Yitao and Ma, Jianzhu and Zou, James and Ermon, Stefano},
  journal={Advances in Neural Information Processing Systems},
  volume={37},
  pages={22370--22417},
  year={2024}
}

@article{agarwal2026cosmos,
  title={Cosmos 3: Omnimodal world models for physical ai},
  author={Agarwal, Niket and Ali, Arslan and Allen, Jon and Antolini, Martin and Aubame, Adeline and Azzolini, Alisson and Bai, Junjie and Bala, Maciej and Balaji, Yogesh and Bapst, Josh and others},
  journal={arXiv preprint arXiv:2606.02800},
  year={2026}
}

@article{liu2026breaking,
  title={Breaking dual bottlenecks: Evolving unified multimodal models into self-adaptive interleaved visual reasoners},
  author={Liu, Qingyang and Gao, Bingjie and Fu, Canmiao and Huang, Zhipeng and Li, Chen and Wang, Feng and Chang, Shuochen and Wang, Shaobo and Wang, Yali and Ye, Keming and others},
  journal={arXiv preprint arXiv:2605.14709},
  year={2026}
}

@inproceedings{heusel2017gan,
  title     = {GANs Trained by a Two Time-Scale Update Rule Converge
 to a Local {Nash} Equilibrium},
  author    = {Heusel, Martin and Ramsauer, Hubert and Unterthiner,
               Thomas and Nessler, Bernhard and Hochreiter, Sepp},
  booktitle = {Advances in Neural Information Processing Systems},
  year      = {2017}
}

@inproceedings{radford2021clip,
  title     = {Learning Transferable Visual Models From Natural               Language Supervision},
  author    = {Radford, Alec and Kim, Jong Wook and Hallacy, Chris and
               others},
  booktitle = {International Conference on Machine Learning},
  year      = {2021}
}

@article{han2026vision,
  title={Vision as Unified Multimodal Generation},
  author={Han, Xiaoyang and Li, Jianhua and Deng, Kewang and Chen, Zukai and Shi, Xuanke and Wang, Sihan and Li, Boxuan and Wang, Linyan and Xie, Siyi and You, Xin and others},
  journal={arXiv preprint arXiv:2607.06560},
  year={2026}
}

@article{oquab2024dinov2,
  title   = {DINOv2: Learning Robust Visual Features without Supervision},
  author  = {Oquab, Maxime and Darcet, Timoth{\'e}e and Moutakanni,
             Theo and Vo, Huy and Szafraniec, Marc and Khalidov,
             Vasil and Fernandez, Pierre and Haziza, Daniel and Massa, Francisco and El-Nouby, Alaaeldin and others},
  journal = {Transactions on Machine Learning Research},
  year    = {2024}
}
\bibliographystyle{iclr2027_conference}

\clearpage
\appendix
\section{Appendix}
\subsection{Extension to Video Generation}
\label{app:video-extension}

In our video experiments, a GPT-5-based planner is used as a required
preprocessing stage. Given a textual prompt $\mathcal{P}$ and a target length
of $F$ frames, it decomposes the prompt into $K$ temporally ordered events and
allocates $n_k$ frames to each event:
\begin{equation}
\Pi(\mathcal{P},F)
=
\left\{
\left(e_k^{\mathrm{v}},n_k\right)
\right\}_{k=1}^{K},
\qquad
\sum_{k=1}^{K}n_k=F .
\label{eq:video-event-planning}
\end{equation}
Let $b_k=\sum_{j=1}^{k}n_j$ with $b_0=0$. The event stage assigned to frame
$t$ is denoted by $\kappa(t)=k$ when
$b_{k-1}<t\leq b_k$.

Let $\mathcal{H}^{\mathrm{v}}_t$ denote the temporal context inherited by
frame $t$. Following the image-slot formulation, we define
\begin{equation}
\begin{aligned}
c_{\mathrm{full}}^{(t)}
&=
\left(\mathcal{H}^{\mathrm{v}}_t,e_{\kappa(t)}^{\mathrm{v}}\right),&
c_{\mathrm{counter}}^{(t)}
&=
\left(\mathcal{H}^{\mathrm{v}}_t,
e_{\varnothing,\kappa(t)}^{\mathrm{v}}\right),&
c_{\mathrm{event}}^{(t)}
&=
\left(\varnothing,e_{\kappa(t)}^{\mathrm{v}}\right),
\end{aligned}
\label{eq:video-conditions}
\end{equation}
where $e_{\varnothing,\kappa(t)}^{\mathrm{v}}$ describes the temporal state
immediately before the event assigned to frame $t$.

For a jointly generated video, let $\mathbf{c}_b$ denote the complete
frame-aligned condition schedule for branch $b$, and let
$v_b^{(t)}=[v_\theta(z_\tau\mid\mathbf{c}_b)]_t$ be its frame-indexed output
at the same spatiotemporal latent state $z_\tau$. The frame-wise CFG update and
the two SCO directions are then
\begin{equation}
\begin{aligned}
v_{\mathrm{ref}}^{(t)}
&=
v_{\mathrm{u}}^{(t)}
+
\omega\left(
v_{\mathrm{f}}^{(t)}-v_{\mathrm{u}}^{(t)}
\right),
&
\delta_{\mathrm{ref}}^{(t)}
&=
v_{\mathrm{ref}}^{(t)}-v_{\mathrm{u}}^{(t)},\\
a_{\mathrm{new}}^{(t)}
&=
v_{\mathrm{f}}^{(t)}-v_{\mathrm{counter}}^{(t)},
&
a_{\mathrm{state}}^{(t)}
&=
v_{\mathrm{f}}^{(t)}-v_{\mathrm{event}}^{(t)} .
\end{aligned}
\label{eq:video-frame-updates}
\end{equation}
Although the CFG scale $\omega$ is shared, these updates remain
frame-dependent because the event assignments and temporal contexts vary
across frames.

The new-event direction captures the motion or physical transition associated
with the assigned event, whereas the state-preserving direction retains
persistent information from the preceding temporal context. Using the same
strength schedule and bounded joint optimization as in
\eqref{eq:strength-scheduling}--\eqref{eq:correction-budgets}, SCO computes
\begin{equation}
\delta^{*(t)}
=
\delta_{\mathrm{ref}}^{(t)}
+
\mu_{\mathrm{new}}^{(t)}a_{\mathrm{new}}^{(t)}
+
\mu_{\mathrm{state}}^{(t)}a_{\mathrm{state}}^{(t)} .
\label{eq:video-correction}
\end{equation}
For jointly generated videos, $v^{(t)}$ denotes the temporal component of the
shared spatiotemporal update rather than an independently sampled frame.
Therefore, the same formulation applies to both diffusion- and
flow-matching-based video generators while preserving their cross-frame
interactions.

\subsection{Dataset details}
\label{dataset}
\paragraph{Interleaved image--text generation.}
We evaluate image--text generation on two challenging benchmarks, OpenING and ISG-Bench. OpenING~\cite{zhou2025opening} is a large-scale benchmark for open-ended interleaved image--text generation. It contains 5,400 human-annotated instances spanning 23 real-world meta-topics and 56 fine-grained tasks. The benchmark covers diverse application domains, including visual explanation, planning, design, education, travel, and creative content generation. Each instance specifies an open-ended multimodal request whose response consists of interleaved textual descriptions and generated images. OpenING therefore evaluates the complete generation process, including the helpfulness of the response, text quality, visual quality, text--image alignment, and coherence across multiple generated images. We use the official benchmark prompts and evaluate the generated responses with the official evaluation protocol.

ISG-Bench~\cite{chen2025interleaved} provides a more structured evaluation of interleaved generation. It contains 1,150 samples organized into 8 scenarios and 21 subtasks. Each sample is associated with an interleaved scene graph that decomposes the input into atomic entities, attributes, relations, and generation requirements. The benchmark evaluates generation at four levels: structural consistency, block-level consistency, image-level correctness, and holistic interleaved response quality. Compared with OpenING, ISG-Bench places greater emphasis on fine-grained language--vision dependencies and allows individual semantic requirements to be verified separately. We use ISG-Bench to measure whether SCO improves the realization of the newly requested event while preserving entities, attributes, spatial relations, and visual style established earlier in the sequence.

\begin{figure}[!t]
\centerline{\includegraphics[width=1.0\linewidth]{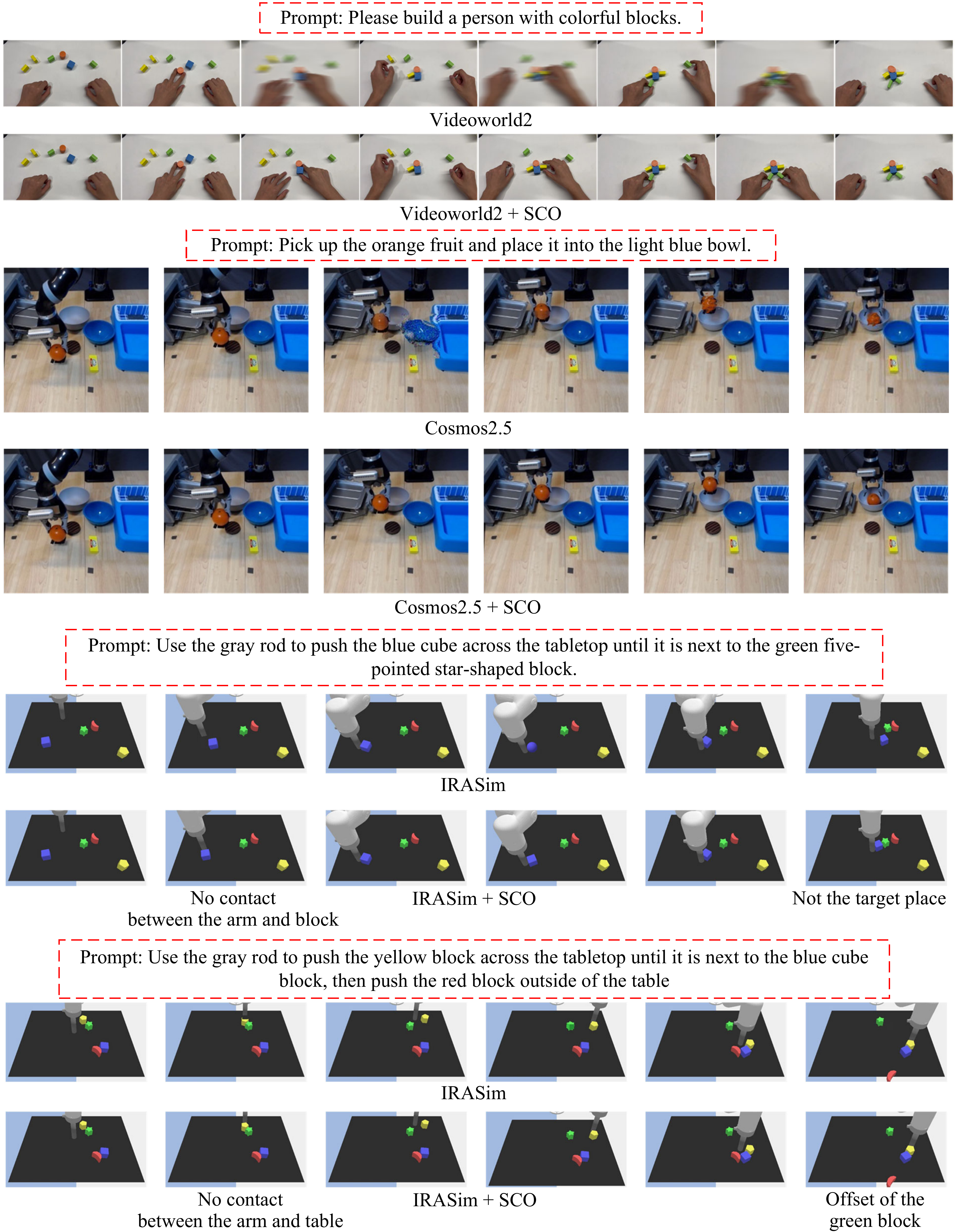}}
\caption{Qualitative results on the effectiveness of SCO when applied to embodiment and handcrafting fields.}
\label{appendix_video_gen}
\end{figure}
\paragraph{Video generation and embodied scenarios.}
To examine whether the proposed constraint extends beyond still-image generation, we additionally conduct experiments on two video-oriented data sources. First, we use the handcraft-making portion of Video-CraftBench, introduced by VideoWorld~2~\cite{ren2026videoworld}. Video-CraftBench consists of first-person videos depicting long-horizon manipulation tasks, including paper-folding and block-building activities. The benchmark contains five handcrafted tasks and approximately 9.5K video clips collected from about seven hours of demonstrations. The tasks involve multiple fine-grained steps, substantial object and background variation, and long temporal horizons, making them suitable for evaluating both event realization and state preservation.

Second, we use a subset of the Interleaved X-Embodiment data released with Interleave-VLA~\cite{fan2026interleave}. Interleaved X-Embodiment is constructed from the Open X-Embodiment collection ~\cite{o2024open} and contains approximately 210K robot-manipulation trajectories. The data-generation pipeline converts text-only manipulation instructions into interleaved image--text instructions by identifying task-relevant objects, extracting corresponding image regions, and verifying the resulting image--text pairs with vision--language models. For the video experiments, we report both event-oriented and preservation-oriented metrics. Event-oriented metrics measure whether the requested manipulation or state transition appears in the generated clip. Preservation-oriented metrics measure identity consistency, object attributes, scene layout, and visual similarity to the conditioning frames. This setup enables us to test whether SCO reduces temporal and cross-frame drift without suppressing the requested action.

We further evaluate SCO on Language-Table~\cite{lynch2023interactive}, a simulated tabletop manipulation benchmark for language-guided visuomotor learning. The environment contains a robot arm operating on a two-dimensional tabletop workspace, where natural-language instructions specify manipulation goals such as moving, pushing, or arranging objects. Each episode consists of visual observations, language annotations, and the corresponding low-dimensional action trajectory. Compared with the image--text benchmarks, Language-Table introduces an embodied setting in which the requested event is grounded in a physical state transition and must remain consistent with the robot trajectory. Following IRASim~\cite{zhu2025irasim}, we formulate the task as trajectory-conditioned video generation. Given an initial observation and a short sequence of robot actions, the model generates the corresponding future video frames.

\begin{figure}[!t]
\centerline{\includegraphics[width=1.0\linewidth]{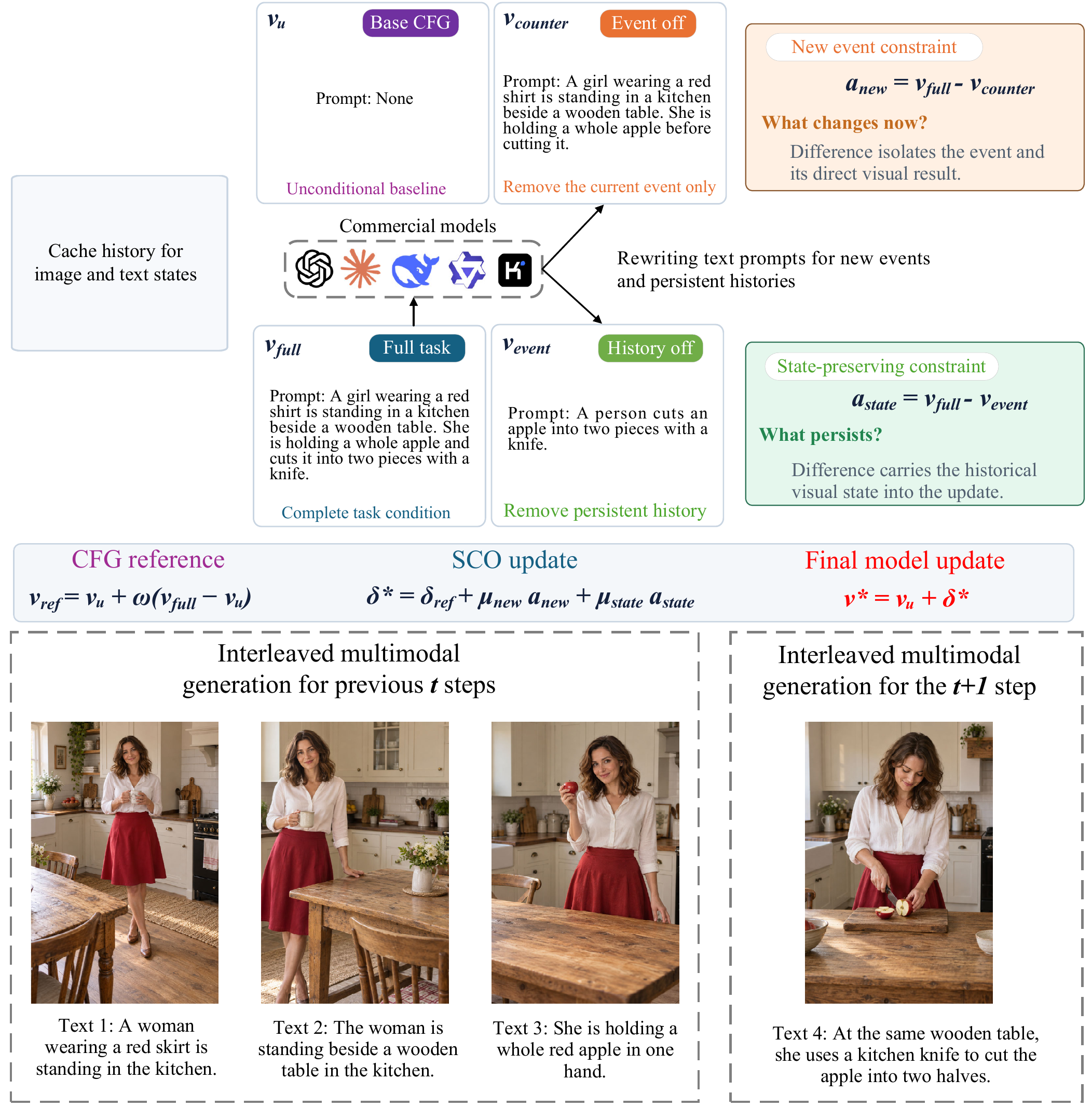}}
\caption{The relationship between the new-event constraint/state-preserving constraint and how to rewrite text prompts.}
\label{appendix_prompt_rewriting}
\end{figure}

\subsection{Details of Evaluation Metrics}
\label{app:metrics}

We evaluate along four complementary dimensions: image generation
quality, cross-modality consistency, temporal coherence, and fact
consistency with physical plausibility. We combine reference-based
automatic metrics with GPT-based qualitative judgments for a holistic
assessment. All GPT-based judgments use GPT-5.6-sol with fixed sampling
settings.

\paragraph{Image Generation Quality:}Fréchet Inception Distance (FID).
FID~\citep{heusel2017gan} quantifies the gap between the distributions
of generated and real images in the feature space of a pre-trained
Inception-V3 network. Let $(\mu_r,\Sigma_r)$ and $(\mu_g,\Sigma_g)$
denote the mean and covariance of features extracted from the real and
generated sets, respectively:
\begin{equation}
\mathrm{FID}
 \;=\;
 \|\mu_r-\mu_g\|_2^{2}
  \;+\;
  \mathrm{Tr}\!\left(
 \Sigma_r+\Sigma_g-2\bigl(\Sigma_r\Sigma_g\bigr)^{1/2}
  \right).
\end{equation}
Lower FID indicates a closer match between the generated and real
distributions.

\paragraph{Image Generation Quality:}CLIP Image Similarity (CLIP-I).
We compute the cosine similarity between CLIP image
embeddings~\citep{radford2021clip} of the prediction and the
ground-truth frame, providing a frame-level measure of perceptual
similarity that complements FID's distributional view.

\paragraph{Cross-Modality Consistency:}CLIP Text--Image Similarity (CLIP-T).
Defined as the cosine similarity between the CLIP text embedding of the
input prompt and the CLIP image embedding of the generated frame,
CLIP-T captures fine-grained semantic alignment between modalities.

\paragraph{Cross-Modality Consistency:}GPT-Judge 1 (GPT-J1).
To overcome the limits of embedding-based similarity, we employ a
multimodal large language model, GPT-5.6-sol, as a
judge. GPT-J1 scores each frame on a 0--1 scale along two axes:
(i) whether the generated image faithfully reflects the textual
description, and (ii) whether the image is factually and logically
plausible.

\paragraph{Temporal Coherence:}Subject Consistency ($C_s$).
We measure the cosine similarity between DINOv2~\citep{oquab2024dinov2}
features of the primary subject region across consecutive frames. The
self-supervised features of DINOv2 are sensitive to high-level semantic
structure, making them well-suited for tracking identity-preserving
subjects.

\paragraph{Temporal Coherence:}Background Consistency ($C_b$).
We compute the cosine similarity between CLIP image embeddings of the
background region across frames to assess low-level scene stability.

\paragraph{Fact Consistency and Physical Plausibility:}GPT-Judge 2 (GPT-J2).
Using the same GPT-5.6-sol backbone as GPT-J1, GPT-J2 provides a holistic
0--1 rating along two axes: (i) \emph{fact consistency} -- whether the
depicted events align with established real-world facts (e.g.,
object identities, causal relations), and (ii) \emph{physical
plausibility} -- whether the implied dynamics obey common-sense physics
(e.g., gravity, rigid-body motion, material interactions).
\paragraph{Remark.}
We use the same prompt template and sampling settings across all baselines. Besides, absolute scores should be compared only within the same GPT-Judge run.

\subsection{Prompt construction for SCO}
Figure~\ref{appendix_prompt_rewriting} illustrates how prompt rewriting supports the two complementary SCO constraints. For each generation round, the full-task prompt combines the persistent visual state retrieved from the preceding multimodal history with the newly requested event. To construct the new-event constraint, a counterfactual prompt retains the subject, objects, and scene established in previous rounds while removing only the current event; comparison with the full condition therefore isolates the visual change associated with that event. To construct the state-preserving constraint, an event-only prompt retains the current action while removing historical details such as subject identity, appearance, and scene configuration; comparison with the full condition consequently identifies the persistent state that should be preserved. In the illustrated example, these rewrites separate cutting the apple from the previously established woman, clothing, kitchen, table, and object states.

\begin{figure}[!t]
\centerline{\includegraphics[width=1.0\linewidth]{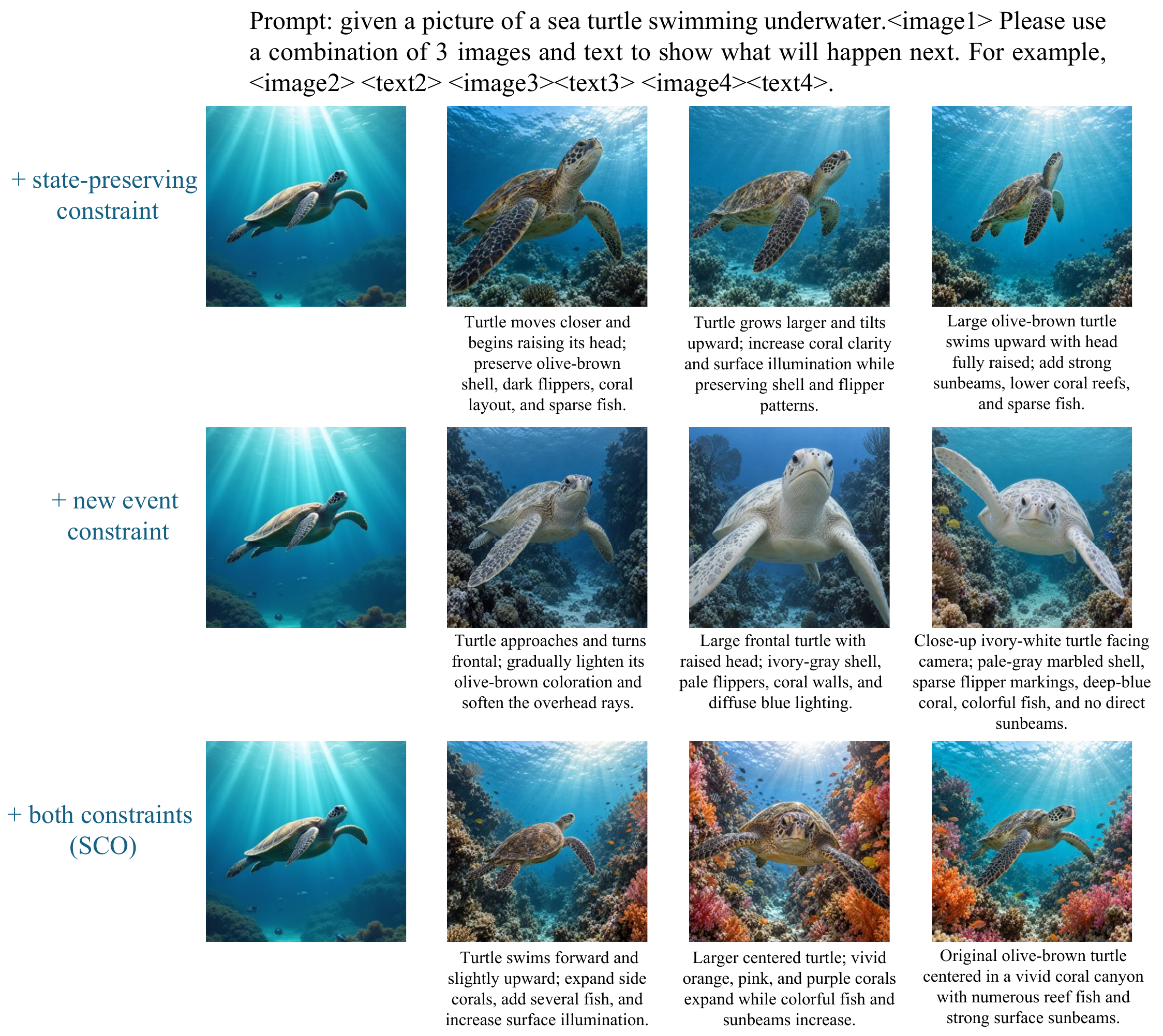}}
\caption{Qualitative results on the interleaved multimodal generation for the storytelling task.}
\label{appendix_storytelling}
\end{figure}

\subsection{Additional qualitative results}
\begin{figure}[!t]
\centerline{\includegraphics[width=0.85\linewidth]{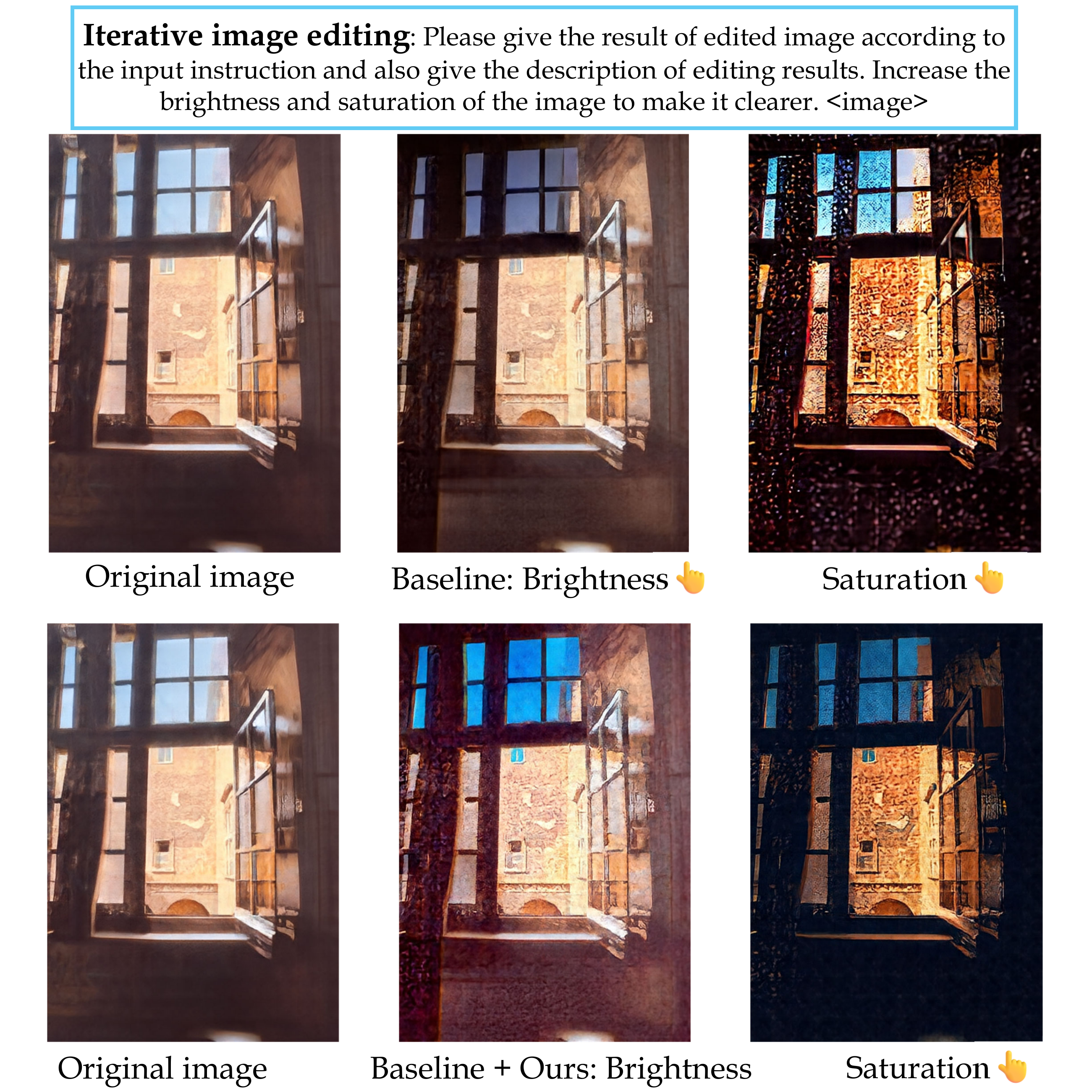}}
\caption{Qualitative results on the task of iterative image editing.}
\label{qualitative_results_image_editing}
\end{figure}

The additional qualitative results further demonstrate the complementary effects of the two SCO constraints. In the storytelling example as shown in Fig.\ref{appendix_storytelling}, the state-preserving constraint maintains persistent visual attributes, such as the sea turtle's olive-brown coloration and characteristic shell and flipper patterns, but produces relatively conservative motion. In contrast, the new-event constraint introduces more pronounced motion and viewpoint changes, while gradually altering the turtle's appearance. By combining both constraints, SCO preserves the turtle's visual identity while producing a richer progression of forward and upward motion, scale variation, and scene evolution. This benefit is more evident in iterative image editing, where each new instruction must be applied without overwriting previously established content. As revealed in Fig.\ref{qualitative_results_image_editing}, compared with the baseline, which accumulates substantial appearance distortions after successive brightness and saturation edits, SCO better preserves the window structure, spatial layout, and overall scene content while incorporating the requested changes. In two scenarios mentioned above, we adopt SenseNova-U1 \cite{diao2026sensenova} as the baseline model.

\begin{figure}[!t]
\centerline{\includegraphics[width=1.0\linewidth]{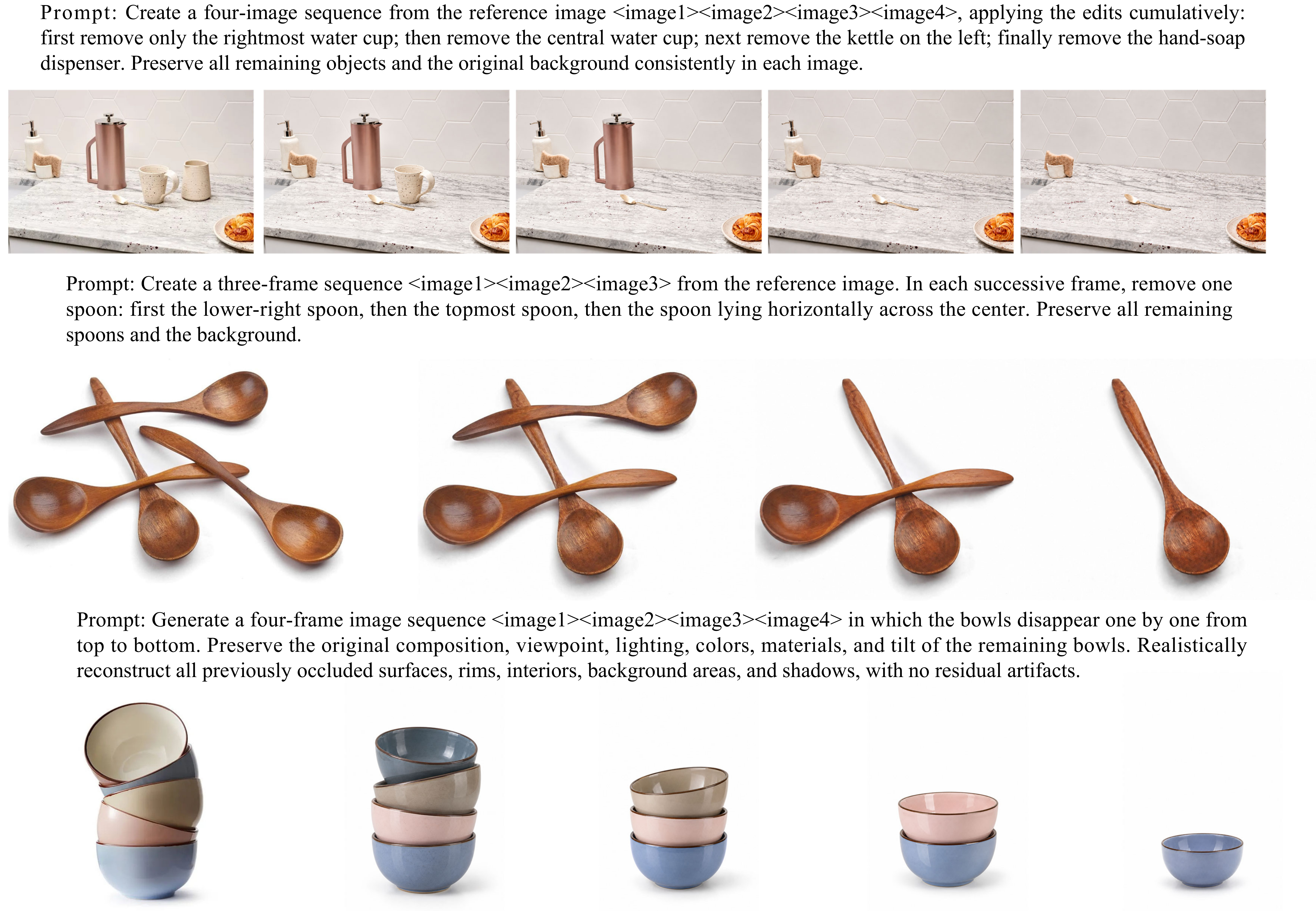}}
\caption{Qualitative results on the interleaved multimodal generation for the sequential object removal task.}
\label{appendix_object_removing}
\end{figure}

Furthermore, in the sequential object removing task, we find that SCO can boost the generative performance of DuoGen\cite{shi2026duogen}. As shown in Fig.\ref{appendix_object_removing}, across scenes containing kitchen objects, wooden spoons, and stacked bowls, the model removes the specified object at each generation round while preserving the identity, appearance, and spatial configuration of the remaining objects. It also maintains the background, viewpoint, lighting, and overall composition across the sequence, while reconstructing regions previously occluded by the removed objects. These examples show that SCO enables DuoGen to follow cumulative removal instructions while maintaining coherent visual states over multiple generation rounds.

\end{document}